\documentclass{article}

\PassOptionsToPackage{numbers, compress}{natbib}

\usepackage[final]{nips_2018}

\usepackage[utf8]{inputenc} 
\usepackage[T1]{fontenc}    
\usepackage{hyperref}       
\usepackage{url}            
\usepackage{booktabs}       
\usepackage{amsfonts}       
\usepackage{nicefrac}       
\usepackage{microtype}      
\usepackage{amsmath}
\usepackage{wrapfig}
\usepackage[font={small}]{caption}
\usepackage{subfigure}
\usepackage{enumitem}
\usepackage{booktabs} 
\usepackage{algorithm}
\usepackage[noend]{algpseudocode}

\usepackage{booktabs} 

\usepackage{url}
\usepackage{boldline}
\usepackage{color}

\usepackage[colorinlistoftodos]{todonotes}

\newcommand{\NSGS}{{NSGS}}
\newcommand{\RProp}{{GRProp}}
\title{Hierarchical Reinforcement Learning for Zero-shot \\ Generalization with Subtask Dependencies}

\author{
  Sungryull Sohn\\
  University of Michigan\\
  \texttt{srsohn@umich.edu} \\
  \And
  Junhyuk Oh\thanks{Now at DeepMind.}\\
  University of Michigan\\
  \texttt{junhyuk@google.com} \\
  \And
  Honglak Lee\\
  Google Brain\\
  University of Michigan\\
  \texttt{honglak@google.com} \\
}

\begin{document}

\maketitle

\newcommand{\fix}{\marginpar{FIX}}
\newcommand{\new}{\marginpar{NEW}}
\newcommand{\mb}{\mathbf}
\newcommand{\tb}{\textbf}
\newcommand{\mbb}{\mathbb}
\newcommand{\mc}{\mathcal}
\newcommand{\wt}{\widetilde}
\newcommand{\tr}{\textrm}

\newcommand{\cutabstractup}{\vspace*{-0.05in}}
\newcommand{\cutabstractdown}{\vspace*{-0.05in}}
\newcommand{\cutsectionup}{\vspace*{-0.05in}}
\newcommand{\cutsectiondown}{\vspace*{-0.08in}}
\newcommand{\cutsubsectionup}{\vspace*{-0.05in}}
\newcommand{\cutsubsectiondown}{\vspace*{-0.08in}}
\newcommand{\cutsubsubsectionup}{\vspace*{-0.05in}}
\newcommand{\cutsubsubsectiondown}{\vspace*{-0.08in}}
\newcommand{\cutcaptionup}{\vspace*{-0.0in}}
\newcommand{\cutcaptiondown}{\vspace*{-0.0in}}
\newcommand{\cutparagraphup}{\vspace{-0.1in}}
\newcommand{\cutparagraphdown}{\vspace*{-0.0in}}
\newcommand{\cutitemizeup}{\vspace{-6pt}}
\newcommand{\cutitemizedown}{\vspace{-6pt}}
\newcommand{\cutparup}{\vspace{-0.03in}}

\cutabstractup
\vspace{-0.2in}
\begin{abstract}
\vspace{-0.1in}
We introduce a new RL problem where the agent is required to generalize to a previously-unseen environment characterized by a subtask graph which describes a set of subtasks and their dependencies.
Unlike existing hierarchical multitask RL approaches that explicitly describe what the agent should do at a high level, our problem only describes properties of subtasks and relationships among them, which requires the agent to perform complex reasoning to find the optimal subtask to execute. 
To solve this problem, we propose a \emph{neural subtask graph solver} (\NSGS{}) which encodes the subtask graph using a recursive neural network embedding. To overcome the difficulty of training, we propose a novel non-parametric gradient-based policy, \emph{graph reward propagation}, to pre-train our \NSGS{} agent and further finetune it through actor-critic method. The experimental results on two 2D visual domains show that our agent can perform complex reasoning to find a near-optimal way of executing the subtask graph and generalize well to the unseen subtask graphs. 
In addition, we compare our agent with a Monte-Carlo tree search (MCTS) method showing that our method is much more efficient than MCTS, and the performance of \NSGS{} can be further improved by combining it with MCTS.
\vspace{-0.1in}
\end{abstract}

\cutabstractdown
\cutsectionup
\section{Introduction}\label{sec:i}
\cutsectiondown
Developing the ability to execute many different tasks depending on given task descriptions and generalize over unseen task descriptions is an important problem for building scalable reinforcement learning (RL) agents. Recently, there have been a few attempts to define and solve different forms of task descriptions such as natural language~\citep{oh2017zero,yu2017deep} or formal language~\citep{denil2017programmable,andreas2017modular}. However, most of the prior works have focused on task descriptions which explicitly specify what the agent should do at a high level, which may not be readily available in real-world applications.

\cutparup
To further motivate the problem, let's consider a scenario in which an agent needs to generalize to a complex novel task by performing a composition of subtasks where the task description and dependencies among subtasks may change depending on the situation. 
For example, a human user could ask a physical household robot to make a meal in an hour. A meal may be served with different combinations of dishes, each of which takes a different amount of cost (e.g., time) and gives a different amount of reward (e.g., user satisfaction) depending on the user preferences. In addition, there can be complex dependencies between subtasks. For example, a bread should be sliced before toasted, or an omelette and an egg sandwich cannot be made together if there is only one egg left. Due to such complex dependencies as well as different rewards and costs, it is often cumbersome for human users to manually provide the optimal sequence of subtasks (e.g., ``fry an egg and toast a bread'').
Instead, the agent should learn to act in the environment by figuring out the optimal sequence of subtasks that gives the maximum reward within a time budget just from properties and dependencies of subtasks.

\cutparup
\begin{figure*}[t]
\vspace{-6pt}
    \centering
   \includegraphics[width=0.99\linewidth]{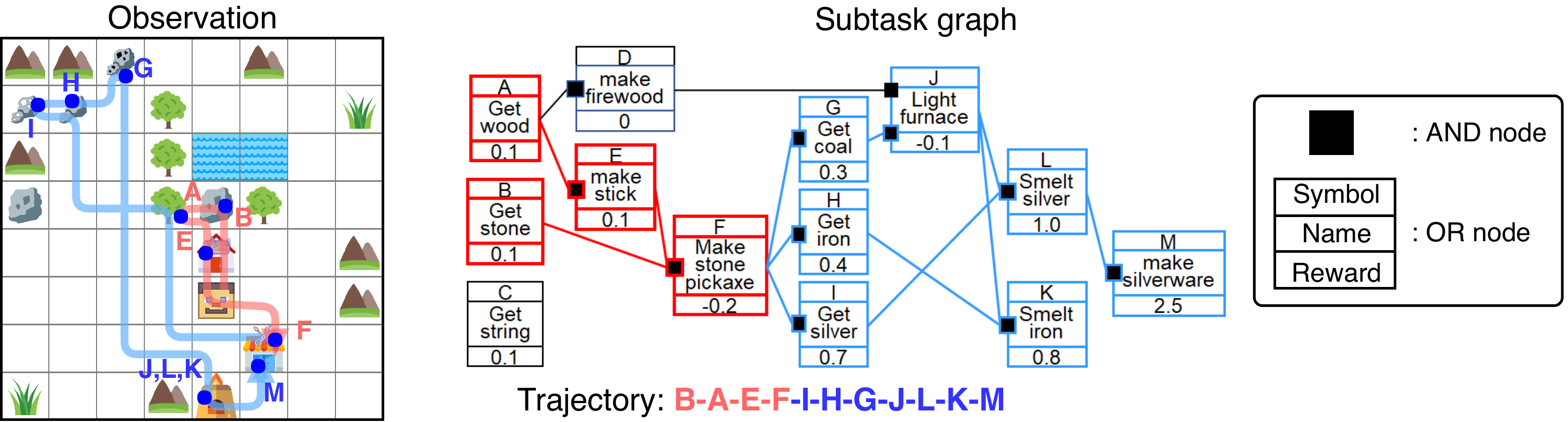}
   \vspace{-0.02in}
   \caption{Example task and our agent's trajectory. The agent is required to execute subtasks in the optimal order to maximize the reward within a time limit. The subtask graph describes subtasks with the corresponding rewards (e.g., subtask L gives 1.0 reward) and dependencies between subtasks through \texttt{AND} and \texttt{OR} nodes. For instance, the agent should first get the firewood (D) \texttt{OR} coal (G) to light a furnace (J). In this example, our agent learned to execute subtask F and its preconditions (shown in red) as soon as possible, since it is a precondition of many subtasks even though it gives a negative reward. After that, the agent mines minerals that require stone pickaxe and craft items (shown in blue) to achieve a high reward.}
   \label{fig:intro}
   \vspace{-15pt}
\end{figure*}
The goal of this paper is to formulate and solve such a problem, which we call \textit{subtask graph execution}, where the agent should execute the given \textit{subtask graph} in an optimal way as illustrated in Figure~\ref{fig:intro}.
A subtask graph consists of subtasks, corresponding rewards, and dependencies among subtasks in logical expression form where it subsumes many existing forms (e.g., sequential instructions~\citep{oh2017zero}). This allows us to define many complex tasks in a principled way and train the agent to find the optimal way of executing such tasks. 
Moreover, we aim to solve the problem without explicit search or simulations so that our method can be more easily applicable to practical real-world scenarios, where real-time performance (i.e., fast decision-making) is required and building the simulation model is extremely challenging.
%

\cutparup
To solve the problem, we propose a new deep RL architecture, called \textit{neural subtask graph solver} (\NSGS{}), which encodes a subtask graph using a recursive-reverse-recursive neural network (R3NN)~\citep{NSPS} to consider the long-term effect of each subtask. 
Still, finding the optimal sequence of subtasks by reflecting the long-term dependencies between subtasks and the context of observation is computationally intractable. Therefore, we found that it is extremely challenging to learn a good policy when it's trained from scratch.
%
%
%
To address the difficulty of learning, we propose to pre-train the \NSGS{} to approximate our novel non-parametric policy called \textit{graph reward propagation policy}. The key idea of the graph reward propagation policy is to construct a differentiable representation of the subtask graph such that taking a gradient over the reward results in propagating reward information between related subtasks, which is used to find a reasonably good subtask to execute.  
After the pre-training, our \NSGS{} architecture is finetuned using the actor-critic method. 

\cutparup
The experimental results on 2D visual domains with diverse subtask graphs show that our agent implicitly performs complex reasoning by taking into account long-term subtask dependencies as well as the cost of executing each subtask from the observation, and it can successfully generalize to unseen and larger subtask graphs. Finally, we show that our method is computationally much more efficient than Monte-Carlo tree search (MCTS) algorithm, and the performance of our \NSGS{} agent can be further improved by combining with MCTS, achieving a near-optimal performance.

\cutparup
Our contributions can be summarized as follows: 
(1) We propose a new challenging RL problem and domain with a richer and more general form of graph-based task descriptions compared to the recent works on multitask RL.
(2) We propose a deep RL architecture that can execute arbitrary \emph{unseen} subtask graphs and observations.
(3) We demonstrate that our method outperforms the state-of-the-art search-based method (e.g., MCTS), which implies that our method can efficiently approximate the solution of an intractable search problem without performing any search. %
(4) We further show that our method can also be used to augment MCTS, which significantly improves the performance of MCTS with a much less amount of simulations. 
\cutsectionup
\section{Related Work}\label{sec:r}
\cutsectiondown
\paragraph{Programmable Agent}
The idea of learning to execute a given program using RL was introduced by programmable hierarchies of abstract machines (PHAMs)~\citep{Parr1997ReinforcementLW,Andre2000ProgrammableRL,Andre2002StateAF}. 
PHAMs specify a partial policy using a set of hierarchical finite state machines, and the agent learns to execute the partial program. A different way of specifying a partial policy was explored in the deep RL framework~\citep{andreas2017modular}. Other approaches used a program as a form of task description rather than a partial policy in the context of multitask RL~\citep{oh2017zero,denil2017programmable}. Our work also aims to build a programmable agent in that we train the agent to execute a given task. However, most of the prior work assumes that the program specifies what to do, and the agent just needs to learn how to do it. In contrast, our work explores a new form of program, called \textit{subtask graph} (see Figure~\ref{fig:intro}), which describes properties of subtasks and dependencies between them, and the agent is required to figure out what to do as well as how to do it. 
\cutparagraphup
\paragraph{Hierarchical Reinforcement Learning}
Many hierarchical RL approaches have been proposed to solve complex decision problems via multiple levels of temporal abstractions~\citep{sutton1999between,dietterich2000hierarchical,precup2000temporal,ghavamzadeh2003hierarchical,Konidaris2007BuildingPO}.
Our work builds upon the prior work in that a high-level controller focuses on finding the optimal subtask, while a low-level controller focuses on executing the given subtask. 
In this work, we focus on how to train the high-level controller for generalizing to novel complex dependencies between subtasks.
\cutparagraphup
\paragraph{Classical Search-Based Planning}
One of the most closely related problems is the planning problem considered in hierarchical task network (HTN) approaches~\citep{sacerdoti1975nonlinear, erol1996hierarchical, erol1994umcp,nau1999shop,castillo2005temporal} in that HTNs also aim to find the optimal way to execute tasks given subtask dependencies. However, they aim to execute a single goal task, while the goal of our problem is to maximize the cumulative reward in RL context. Thus, the agent in our problem not only needs to consider dependencies among subtasks but also needs to infer the cost from the observation and deal with stochasticity of the environment. These additional challenges make it difficult to apply such classical planning methods to solve our problem. 
\cutparagraphup
\paragraph{Motion Planning}
Another related problem to our subtask graph execution problem is motion planning (MP) problem~\citep{asano1985visibility,canny1985voronoi,canny1987new,faverjon1987local,keil1985minimum}. MP problem is often mapped to a graph, and reduced to a graph search problem. However, different from our problem, the MP approaches aim to find an optimal path to the goal in the graph while avoiding obstacles similar to HTN approaches.
\cutsectionup
\section{Problem Definition}\label{sec:p}
\cutsectiondown
\subsection{Preliminary: Multitask Reinforcement Learning and Zero-Shot Generalization}
\cutsubsectiondown
We consider an agent presented with a task drawn from some distribution as in~\cite{andreas2017modular, da2012learning}. 
We model each task as Markov Decision Process (MDP). Let $G\in\mc{G}$ be a task parameter available to agent drawn from a distribution $P(G)$ where $G$ defines the task and $\mc{G}$ is a set of all possible task parameters. The goal is to maximize the expected reward over the whole distribution of MDPs: $\int{P(G)J(\pi,G)dG}$, where $J(\pi, G)= \mathbb{E}_\pi[\sum^{T}_{t=0}{\gamma^{t}r_t}]$ is the expected return of the policy $\pi$ given a task defined by $G$, $\gamma$ is a discount factor, $\pi:\mc{S\times G}\rightarrow \mc{A}$ is a multitask policy that we aim to learn, and $r_t$ is the reward at time step $t$. We consider a zero-shot generalization where only a subset of tasks $\mc{G}_{train}\subset \mc{G}$ is available to agent during training, and the agent is required to generalize over a set of unseen tasks $\mc{G}_{test}\subset \mc{G}$ for evaluation, where $\mc{G}_{test} \cap \mc{G}_{train}=\phi$.\cutsubsectionup
\subsection{Subtask Graph Execution Problem}\label{sec:sge}
\cutsubsectiondown
The \textit{subtask graph execution} problem is a multitask RL problem with a specific form of task parameter $G$ called \textit{subtask graph}. 
Figure~\ref{fig:intro} illustrates an example subtask graph and environment. 
The task of our problem is to execute given $N$ subtasks in an optimal order to maximize reward within a time budget, where there are complex dependencies between subtasks defined by the subtask graph. We assume that the agent has learned a set of \textit{options} ($\mc{O}$)~\cite{precup2000temporal, stolle2002learning, sutton1999between} that performs subtasks by executing one or more primitive actions.
\cutparagraphup
\paragraph{Subtask Graph and Environment} We define the terminologies as follows:
\cutitemizeup
\begin{itemize}[leftmargin=*]
\setlength\itemsep{-0em}
\item{\tb{Precondition}}: A \textit{precondition} of subtask is defined as a logical expression of subtasks in sum-of-products (SoP) form where multiple \texttt{AND} terms are combined with an \texttt{OR} term (e.g., the precondition of subtask J in Figure~\ref{fig:intro} is \texttt{OR}(\texttt{AND}(\texttt{D}), \texttt{AND}(\texttt{G})).
\item{\tb{Eligibility vector}}: $\mb{e}_{t} =[e_t^1,\ldots,e_t^N]$ where $e^i_t=1$ if subtask $i$ is \textit{eligible} (i.e., the precondition of subtask is satisfied and it has never been executed by the agent) at time $t$, and $0$ otherwise.
\item{\tb{Completion vector}}: $\mb{x}_{t}=[x_t^1,\ldots,x_t^N]$ where $x_{t}^{i}=1$ if subtask $i$ has been executed by the agent while it is eligible, and $0$ otherwise.
\item{\tb{Subtask reward vector}}: $\mb{r}=[r^1,\ldots,r^N]$ specifies the reward for executing each subtask.
\item{\tb{Reward}}: $r_t=r^i$ if the agent executes the subtask $i$ while it is eligible, and $r_t=0$ otherwise.
\item{\tb{Time budget}}: $step_t \in \mathbb{R} $ is the remaining time-steps until episode termination.
\item{\tb{Observation}}: $\mb{obs}_t\in\mbb{R}^{H\times W\times C}$ is a visual observation at time $t$ as illustrated in Figure~\ref{fig:intro}.
\end{itemize}
\cutitemizedown
To summarize, a subtask graph $G$ defines $N$ subtasks with corresponding rewards $\mb{r}$ and the preconditions.
The state input at time $t$ consists of $\mb{s}_t = \{ \mb{obs}_{t},\mb{x}_t,\mb{e}_{t},step_t \}$.
The goal is to find a policy $\pi:\mb{s}_{t},G \mapsto \mb{o}_t$ which maps the given context of the environment to an \textit{option} ($\mb{o}_t \in \mc{O}$). 

\cutparagraphup
\paragraph{Challenges}
Our problem is challenging due to the following aspects:
\cutitemizeup
\begin{itemize}[leftmargin=*]
\setlength\itemsep{-0em}
\item{\tb{Generalization}}: Only a subset of subtask graphs ($\mathcal{G}_{train}$) is available during training, but the agent is required to execute previously unseen and larger subtask graphs ($\mathcal{G}_{test}$).
\item{\tb{Complex reasoning}}: 
The agent needs to infer the long-term effect of executing individual subtasks in terms of reward and cost (e.g., time) and find the optimal sequence of subtasks to execute without any explicit supervision or simulation-based search.
We note that it may not be easy even for humans to find the solution without explicit search due to the exponentially large solution space.
%

\item{\tb{Stochasticity}}: The outcome of subtask execution is stochastic in our setting (for example, some objects are randomly moving). Therefore, the agent needs to consider the expected outcome when deciding which subtask to execute.
\end{itemize}
\cutsectionup
\section{Method}\label{sec:m}
\cutsectiondown
 \begin{figure*}
    \centering
   \includegraphics[width=0.75\linewidth]{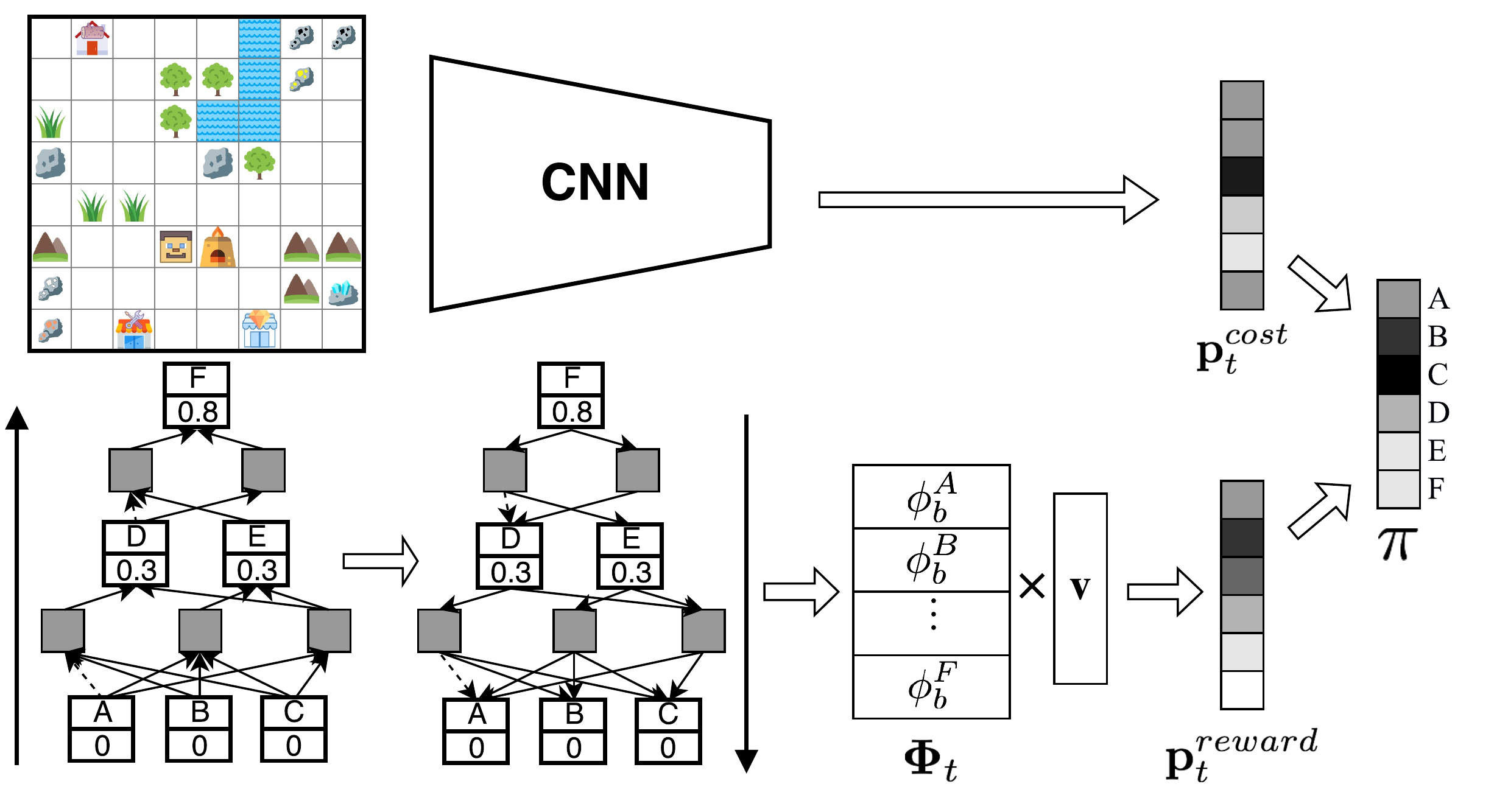}
   \vspace{-10pt}
   \caption{Neural subtask graph solver architecture. The task module encodes subtask graph through a bottom-up and top-down process, and outputs the reward score $\textbf{p}^{reward}_t$. The observation module encodes observation using CNN and outputs the cost score $\textbf{p}^{cost}_t$. The final policy is a softmax policy over the sum of two scores.}
   \label{fig:NSS}
   \vspace{-15pt}
\end{figure*}
Our \textit{neural subtask graph solver} (\NSGS{}) is a neural network which consists of a \textit{task module} and an \textit{observation module} 
as shown in Figure~\ref{fig:NSS}. The task module encodes the precondition of each subtask via bottom-up process and propagates the information about future subtasks and rewards to preceding subtasks (i.e., pre-conditions) via the top-down process. The observation module learns the correspondence between a subtask and its target object, and the relation between the locations of objects in the observation and the time cost.
However, due to the aforementioned challenge (i.e., \textit{complex reasoning}) in Section~\ref{sec:sge}, learning to execute the subtask graph only from the reward is extremely challenging.
To facilitate the learning, we propose \textit{graph reward propagation policy} (\RProp{}), a non-parametric policy that propagates the reward information between related subtasks to model their dependencies. Since our \RProp{} acts as a good initial policy, we train the \NSGS{} to approximate the \RProp{} policy through policy distillation~\citep{rusu2015policy,Parisotto2015ActorMimicDM}, and finetune it through actor-critic method with generalized advantage estimation (GAE)~\citep{schulman2015high} to maximize the reward. Section~\ref{sec:nts} describes the \NSGS{} architecture, and Section~\ref{sec:te} describes how to construct the \RProp{} policy.
 \cutsubsectionup
 \subsection{Neural Subtask Graph Solver} \label{sec:nts}
 \cutsubsectiondown
\paragraph{Task Module}
Given a subtask graph $G$, the remaining time steps $step_t\in\mbb{R}$, an eligibility vector $\mb{e}_t$ and a completion vector $\mb{x}_t$, we compute a context embedding using recursive-reverse-recursive neural network (R3NN)~\cite{NSPS} as follows:
\begin{align}
 \phi^{i}_{bot,o} & = b_{\theta_o}\left(x^{i}_t, e^{i}_t, step_t, \sum_{j \in Child_i}\phi^{j}_{bot,a} \right),&
 \phi^{j}_{bot,a} & = b_{\theta_a}\left(\sum_{k \in Child_j}\left[\phi^{k}_{bot,o},w_{+}^{j,k}\right]\right),\label{eq:rcnn1}\\
 \phi^{i}_{top,o} & = t_{\theta_o}\left(\phi^{i}_{bot,o}, r^i, \sum_{j \in Par_i}\left[\phi^{j}_{top,a},w_{+}^{i,j}\right]\right), &
 \phi^{j}_{top,a} & = t_{\theta_a}\left( \phi^{j}_{bot,a}, \sum_{k \in Par_j}\phi^{k}_{top,o}\right),\label{eq:rcnn2}
\end{align}
where $[\cdot]$ is a concatenation operator, $b_{\theta}, t_{\theta}$ are the bottom-up and top-down encoding function, $\phi^{i}_{bot,a},\  \phi^{i}_{top,a}$ are the bottom-up and top-down embedding of $i$-th \texttt{AND} node respectively, and $\phi^{i}_{bot,o},\ \phi^{i}_{top,o}$ are the bottom-up and top-down embedding of $i$-th \texttt{OR} node respectively (see Appendix for the detail). The $w_{+}^{i,j},\ Child_i$, and $Parent_i$ specifies the connections in the subtask graph $G$. 
Specifically, $w_{+}^{i,j}=1$ if $j$-th \texttt{OR} node and $i$-th \texttt{AND} node are connected without \texttt{NOT} operation, $-1$ if there is \texttt{NOT} connection and $0$ if not connected, and $Child_i, Parent_i$ represent a set of $i$-th node's children and parents respectively.
The embeddings are transformed to reward scores via: $\mb{p}^{reward}_t = \boldsymbol{\Phi}_{top}^{\top}\mb{v},$
where $\boldsymbol{\Phi}_{top} = [\phi_{top,o}^1,\ldots,\phi_{top,o}^N]\in \mathbb{R}^{E \times N}$, $E$ is the dimension of the top-down embedding of \texttt{OR} node, and $\mb{v}\in \mathbb{R}^{E}$ is a weight vector for reward scoring. 

\cutparagraphup
\paragraph{Observation Module}
The observation module encodes the input observation $\mb{s}_{t}$ using a convolutional neural network (CNN) and outputs a cost score:
 \begin{align}
     \mb{p}^{cost}_t =  \tr{CNN}(\mb{s}_{t}, step_t).
 \end{align}
 where $step_t$ is the number of remaining time steps. An ideal observation module would learn to estimate high score for a subtask if the target object is close to the agent because it would require less cost (i.e., time). Also, if the expected number of step required to execute a subtask is larger than the remaining step, ideal agent would assign low score. The \NSGS{} policy is a softmax policy: 
\begin{align}
  \pi(\mb{o}_{t}|\mb{s}_{t},\mb{G},\mb{x}_t,\mb{e}_{t},step_t) = \text{Softmax}( \mb{p}^{reward}_t+\mb{p}^{cost}_t ),
\end{align}
which adds reward scores and cost scores. 

\cutsubsectionup
\subsection{Graph Reward Propagation Policy: Pre-training Neural Subtask Graph Solver} \label{sec:te}
\cutsubsectiondown
Intuitively, the graph reward propagation policy is designed to put high probabilities over subtasks that are likely to maximize the sum of \textit{modified and smoothed} reward $\widetilde{U}_t$ at time $t$, which will be defined in Eq.~\ref{eq:cr4}.
  Let $\mb{x}_t$ be a completion vector and $\mb{r}$ be a subtask reward vector (see Section~\ref{sec:p} for definitions). Then, the sum of reward until time-step $t$ is given as:
 \begin{align}
    U_t &= \mb{r}^{T} \mb{x}_{t}. \label{eq:cr}
 \end{align}
We first modify the reward formulation such that it gives a half of subtask reward for satisfying the preconditions and the rest for executing the subtask to encourage the agent to satisfy the precondition of a subtask with a large reward:
 \begin{align}
    \widehat{U}_t &= \mb{r}^{T} (\mb{x}_{t}+\mb{e}_t)/2. \label{eq:cr2}
 \end{align}
Let $y_{AND}^j$ be the output of $j$-th \texttt{AND} node. The eligibility vector $\textbf{e}_t$ can be computed from the subtask graph $G$ and $\mb{x}_t$ as follows:
\begin{align}
  e_{t}^{i} = \underset{j\in Child_i}{\text{OR}} \left( y^{j}_{AND}\right),\quad
  y^{j}_{AND} = \underset{k\in Child_j}{\text{AND}} \left( \widehat{x}_{t}^{j,k}\right),\quad
  \widehat{x}_{t}^{j,k}= x_t^k w^{j,k} + (1-x_t^k)(1-w^{j,k}),
  \label{eq:xhat}
\end{align}
\begin{wrapfigure}{r}{0.35\textwidth}
\vspace{-16pt}
  \begin{center}
    \includegraphics[width=0.9\linewidth]{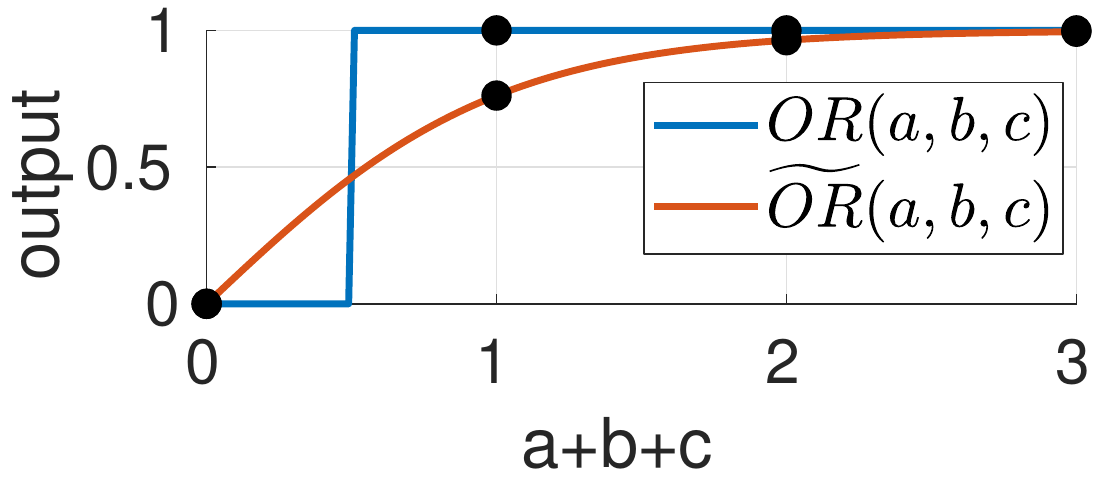}\\
    \includegraphics[width=0.9\linewidth]{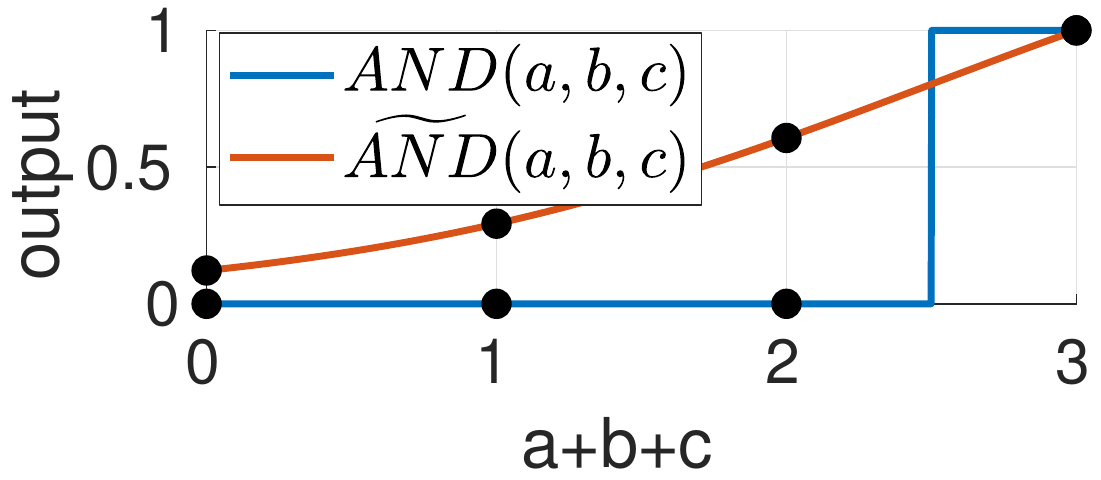}
  \vspace{-8pt}
  \caption{Visualization of \texttt{OR}, $\wt{\rm{\texttt{OR}}}$, \texttt{AND}, and $\wt{\rm{\texttt{AND}}}$ operations with three inputs (a,b,c). These smoothed functions are defined to handle arbitrary number of operands (see Appendix).}\label{fig:activation}
  \end{center}
\vspace{-10pt}
\end{wrapfigure}
where $w^{j, k}=0$ if there is a \texttt{NOT} connection between $j$-th node and $k$-th node, otherwise $w^{j,k}=1$. Intuitively, $\widehat{x}_{t}^{j,k}=1$ when $k$-th node does not violate the precondition of $j$-th node.
Note that $\tilde{U}_t$ is not differentiable with respect to $\mb{x}_t$ because AND$(\cdot)$ and OR$(\cdot)$ are not differentiable. To derive our graph reward propagation policy, we propose to substitute AND$(\cdot)$ and OR$(\cdot)$ functions with ``smoothed'' functions $\wt{ \tr{AND}}$ and $\wt{ \tr{OR} }$ as follows:
  \begin{align}
\wt{e}_{t}^{i} = \underset{j\in Child_i}{\wt{\text{OR}}} \left( \wt{y}^{j}_{AND}\right),\quad
 \wt{y}^{j}_{AND} = \underset{k\in Child_j}{\wt{\text{AND}}} \left( \widehat{x}_{t}^{j,k}\right),
\end{align}
where $\wt{ \tr{AND}}$ and $\wt{ \tr{OR} }$ were implemented as scaled sigmoid and tanh functions as illustrated by Figure~\ref{fig:activation} (see Appendix for details). 
With the smoothed operations, the sum of smoothed and modified reward is given as:
 \begin{align}
    \widetilde{U}_t &= \mb{r}^{T} (\mb{x}_{t}+\wt{\mb{e}}_t)/2. \label{eq:cr4}
 \end{align}
 Finally, the graph reward propagation policy is a softmax  policy, 
 \begin{align}
 \pi(\mb{o}_{t}|G,\mb{x}_t) =  \text{Softmax}\left( \nabla_{\mb{x}_{t}} \widetilde{U}_{t} \right)
 =\text{Softmax}\left( \frac{1}{2}\mb{r}^{T}+\frac{1}{2}\mb{r}^{T}\nabla_{\mb{x}_{t}} \wt{\mb{e}}_{t} \right),
 \end{align}
 that is the softmax of the gradient of $\widetilde{U}_t$ with respect to $\mb{x}_t$.

 \cutsubsectionup
  \subsection{Policy Optimization} \label{sec:train}
\cutsubsectiondown
The \NSGS{} is first trained through policy distillation by minimizing the KL divergence between \NSGS{} and teacher policy (\RProp{}) as follows:
\begin{align}
    \nabla_{\theta} \mathcal{L}_{1} = \mathbb{E}_{G \sim \mathcal{G}_{train}}\left[\mathbb{E}_{s \sim \pi^{G}_{\theta}}\left[\nabla_{\theta} D_{KL}\left(\pi_T^{G} || \pi^{G}_{\theta} \right)  \right] \right],
\end{align}
where $\theta$ is the parameter of \NSGS{}, $\pi^{G}_{\theta}$ is the simplified notation of \NSGS{} policy with subtask graph $G$, $\pi_T^{G}$ is the simplified notation of teacher (\RProp{}) policy with subtask graph $G$, $D_{KL}$ is KL divergence, 
and $\mathcal{G}_{train}$ is the training set of subtask graphs.
 After policy distillation, we finetune \NSGS{} agent in an end-to-end manner using actor-critic method with GAE~\citep{schulman2015high} as follows:
\begin{align}
\nabla_{\theta} \mathcal{L}_{2} &= \mathbb{E}_{G\sim \mathcal{G}_{train}}\left[\mathbb{E}_{s \sim \pi^{G}_{\theta}} \left[ -\nabla_{\theta}\log\pi^{G}_{\theta}\sum^{\infty}_{l=0} \left(\prod_{n=0}^{l-1}{(\gamma\lambda)^{k_n}}\right) \delta_{t+l} \right] \right],\label{eq:multi-gradient}\\
\delta_t &= r_t + \gamma^{k_t} V^{\pi}_{\theta'}(\mb{s}_{t+1},G)  - V^{\pi}_{\theta'}(\mb{s}_{t},G)\label{eq:deltat},
\end{align} 
where $k_t$ is the duration of option $\mb{o}_t$, $\gamma$ is a discount factor, $\lambda \in \left[0, 1\right]$ is a weight for balancing between bias and variance of the advantage estimation, and $V_{\theta'}^{\pi}$ is the critic network parameterized by $\theta'$. 
During training, we update the critic network to minimize $\mathbb{E}\left[ \left(R_t - V^{\pi}_{\theta'}(\mb{s}_{t},G) \right)^2 \right]$, where $R_t$ is the discounted cumulative reward at time $t$. The complete procedure for training our \NSGS{} agent is summarized in Algorithm~\ref{alg:train}. We used $\eta_d$=1e-4, $\eta_c$=3e-6 for distillation and $\eta_{ac}$=1e-6, $\eta_c$=3e-7 for fine-tuning in the experiment.

\vspace{-5pt}
\begin{algorithm}
\caption{Policy optimization}\label{alg:train}
\begin{algorithmic}[1]
\For{ iteration $n$ }
\State Sample $G\sim \mc{G}_{train}$
\State $\mc{D} = \{ (\mb{s}_t, \mb{o}_t, r_t, R_t, step_t),\ldots \}\sim \pi_{\theta}^{G}$ \Comment{do rollout}
\State $\theta' \gets \theta' + \eta_{c} \sum_{\mc{D}}{\left( \nabla_{\theta'}V^{\pi}_{\theta'}(\mb{s}_{t},G) \right)(R_t - V^{\pi}_{\theta'}(\mb{s}_{t},G) ) }$\Comment{update critic}
\If{ distillation }
\State $\theta \gets \theta + \eta_{d} \sum_{\mc{D}}{ \nabla_{\theta} D_{KL}\left(\pi_T^{G} || \pi^{G}_{\theta} \right) }$ \Comment{update policy}
\ElsIf{fine-tuning}
\State Compute $\delta_t$ from Eq.~\ref{eq:deltat} for all $t$
\State $\theta \gets \theta + \eta_{ac} \sum_{\mc{D}}{ \nabla_{\theta}\log\pi_{\theta}^{G}\sum^{\infty}_{l=0}{ \left(\prod_{n=0}^{l-1}{(\gamma\lambda)^{k_n}}\right) \delta_{t+l}}}$ \Comment{update policy}
\EndIf
\EndFor
\end{algorithmic}
\end{algorithm}
\vspace{-5pt}
\cutsectionup
\section{Experiment}\label{sec:e}
\cutsectiondown
In the experiment, we investigated the following research questions: 1) Does \RProp{} outperform other heuristic baselines (e.g., greedy policy, etc.)? 2) Can \NSGS{} deal with complex subtask dependencies, delayed reward, and the stochasticity of the environment? 3) Can \NSGS{} generalize to unseen subtask graphs? 4) How does \NSGS{} perform compared to MCTS? 5) Can \NSGS{} be used to improve MCTS?
\cutsubsectionup
\subsection{Environment}
\cutsubsectiondown
We evaluated the performance of our agents on two domains: \textbf{Mining} and \textbf{Playground} that are developed based on MazeBase~\citep{sukhbaatar2015mazebase}\footnote{The code is available on https://github.com/srsohn/subtask-graph-execution}. 
We used a pre-trained subtask executer for each domain. The episode length (time budget) was randomly set for each episode in a range such that \RProp{} agent executes $60\%-80\%$ of subtasks on average. The subtasks in the higher layer in subtask graph are designed to give larger reward (see Appendix for details).

\tb{Mining} domain is inspired by Minecraft (see Figures~\ref{fig:intro} and~\ref{fig:qual}). 
The agent may pickup raw materials in the world, and use it to craft different items on different craft stations. There are two forms of preconditions: 1) an item may be an ingredient for building other items (e.g., stick and stone are ingredients of stone pickaxe), and 2) some tools are required to pick up some objects (e.g., agent need stone pickaxe to mine iron ore). The agent can use the item multiple times after picking it once. The set of subtasks and preconditions are hand-coded based on the crafting recipes in Minecraft, and used as a template to generate 640 random subtask graphs. We used 200 for training and 440 for testing.

\tb{Playground} is a more flexible and challenging domain (see Figure~\ref{fig:qual2}).
The subtask graph in Playground was randomly generated, hence its precondition can be any logical expression and the reward may be delayed. Some of the objects randomly move, which makes the environment stochastic. The agent was trained on small subtask graphs, while evaluated on much larger subtask graphs (See Table~\ref{tab:depth}). The set of subtasks is $\mathcal{O}=\mathcal{A}_{int} \times \mathcal{X}$, where $\mathcal{A}_{int}$ is a set of primitive actions to interact with objects, and $\mathcal{X}$ is a set of all types of interactive objects in the domain. We randomly generated 500 graphs for training and 2,000 graphs for testing. Note that the task in playground domain subsumes many other hierarchical RL domains such as Taxi~\citep{taxi}, Minecraft~\citep{oh2017zero} and XWORLD~\citep{yu2017deep}.
In addition, we added the following components into subtask graphs to make the task more challenging:
\vspace{-5pt}
\begin{itemize}[leftmargin=*]
\setlength\itemsep{-0.05em}
     \item {Distractor subtask}: A subtask with only \texttt{NOT} connection to parent nodes in the subtask graph. Executing this subtask may give an immediate reward, but it may make other subtasks ineligible.
     \item {Delayed reward}: Agent receives no reward from subtasks in the lower layers, but it should execute some of them to make higher-level subtasks eligible (see Appendix for fully-delayed reward case).
 \end{itemize}
\vspace{-5pt}
\cutsubsectionup
\subsection{Agents}
\cutsubsectiondown
\begin{wraptable}{r}{7.5cm}
\vspace{-14pt}
  \centering
  \small
      \begin{tabular}{|r|c|c|c|c|c|}
      \hlineB{2}
            \multicolumn{6}{|c|}{Subtask Graph Setting}\\ \hlineB{2}
           &\multicolumn{4}{c|}{Playground}    &Mining\\ \hline
         Task   &\tb{D1}    &\tb{D2}    &\tb{D3}    &\tb{D4}&\tb{Eval}\\ \hline
        Depth   &   4   &   4   &   5   &   6  & 4-10 \\ 
        Subtask &   13  &   15  &   16  &   16  & 10-26 \\\hline
        \multicolumn{6}{c}{ }\vspace{-5pt}\\ \hlineB{2}
            \multicolumn{6}{|c|}{Zero-Shot Performance} \\\hlineB{2}
            &\multicolumn{4}{c|}{Playground}    &Mining\\ \hline
         Task   &\tb{D1}    &\tb{D2}    &\tb{D3}    &\tb{D4}&\tb{Eval}\\ \hline
\NSGS{} (Ours)  &\tb{.820}&\tb{.785} &\tb{.715} &\tb{.527}  &\tb{8.19}\\
\RProp{} (Ours)& .721      & .682  & .623      &  .424     &6.16 \\
Greedy      & .164      & .144  & .178      &  .228     &3.39 \\
Random      & 0         & 0     & 0         &  0        &2.79 \\\hlineB{2}
            \multicolumn{6}{|c|}{Adaptation Performance} \\\hlineB{2}
            &\multicolumn{4}{c|}{Playground}    &Mining\\ \hline
         Task   &\tb{D1}    &\tb{D2}    &\tb{D3}    &\tb{D4}&\tb{Eval}\\ \hline
            
\NSGS{} (Ours)  &\tb{.828}  &\tb{.797}  &\tb{.733}  &\tb{.552}  &\tb{8.58} \\
Independent &.346       &.296       &.193       &.188       &   3.89 \\
      \hline
      \end{tabular}
      \vspace{-6pt}
      \caption{Generalization performance on unseen and larger subtask graphs. (Playground) The subtask graphs in \tb{D1} have the same graph structure as training set, but the graph was unseen. The subtask graphs in \tb{D2}, \tb{D3}, and \tb{D4} have (unseen) larger graph structures. (Mining) The subtask graphs in \tb{Eval} are unseen during training. \NSGS{} outperforms other compared agents on all the task and domain.}
  \label{tab:depth}
  \vspace{-20pt}
\end{wraptable}
We evaluated the following policies:
\vspace{-5pt}
\begin{itemize}[leftmargin=*]
\setlength\itemsep{-0.05em}
\item{\textbf{Random}} policy executes any eligible subtask.
\item{\textbf{Greedy}} policy executes the eligible subtask with the largest reward.
\item{\textbf{Optimal}} policy is computed from exhaustive search on \textit{eligible} subtasks.
\item{\textbf{\RProp{}}} (Ours) is graph reward propagation policy.
\item{\textbf{\NSGS{}}} (Ours) is distilled from \RProp{} policy and finetuned with actor-critic.
\item{\textbf{Independent}} is an LSTM-based baseline trained on each subtask graph independently, similar to Independent model in~\cite{andreas2017modular}. It takes the same set of input as \NSGS{} except the subtask graph.
\end{itemize}
\vspace{-5pt}
To our best knowledge, existing work on hierarchical RL cannot directly address our problem with a subtask graph input. Instead, we evaluated an instance of hierarchical RL method (\tb{Independent} agent) in \tb{adaptation} setting, as discussed in Section~\ref{sec:exp-gene}.
\cutsubsectionup
\subsection{Quantitative Result} 
\cutsubsectiondown
\label{sec:exp-gene}
\begin{wrapfigure}{r}{7.5cm}
\vspace{-10pt}
\centering
  \includegraphics[width=0.49\linewidth]{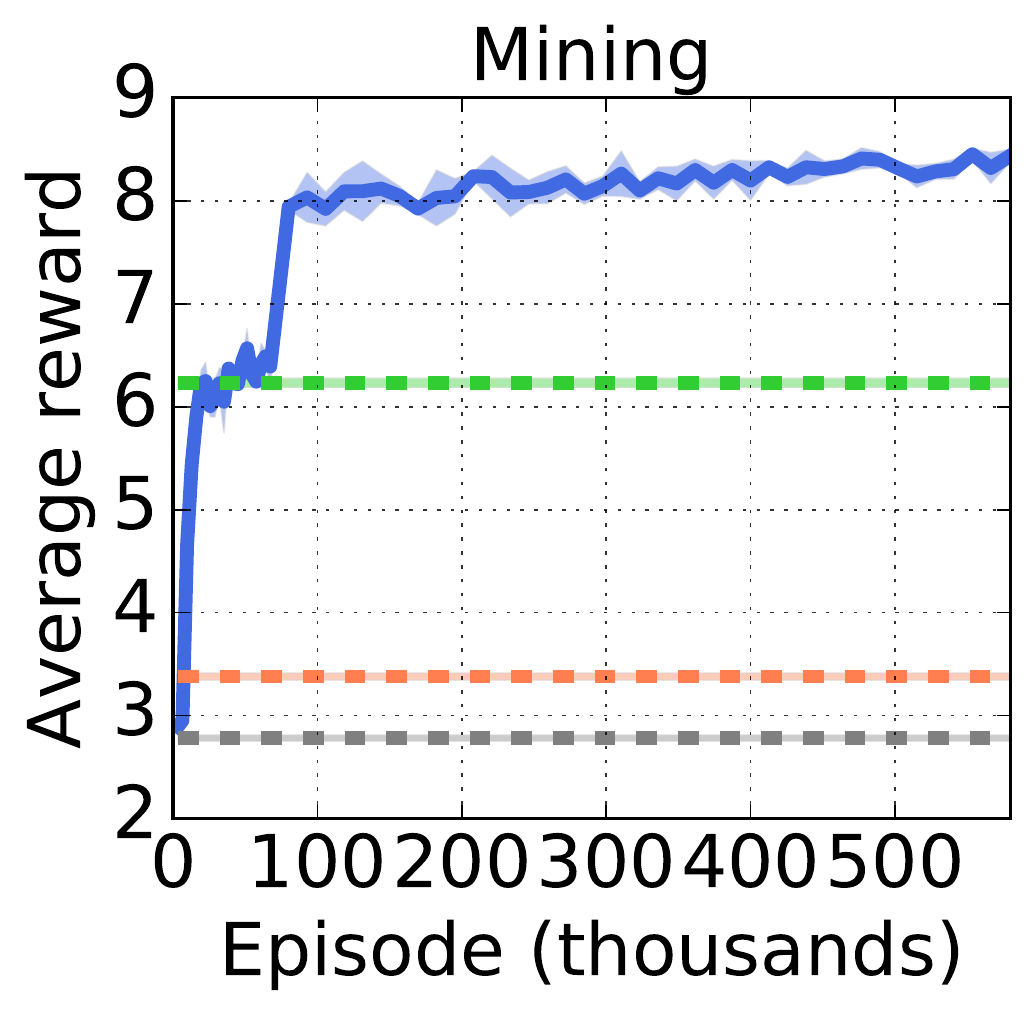}
  \includegraphics[width=0.49\linewidth]{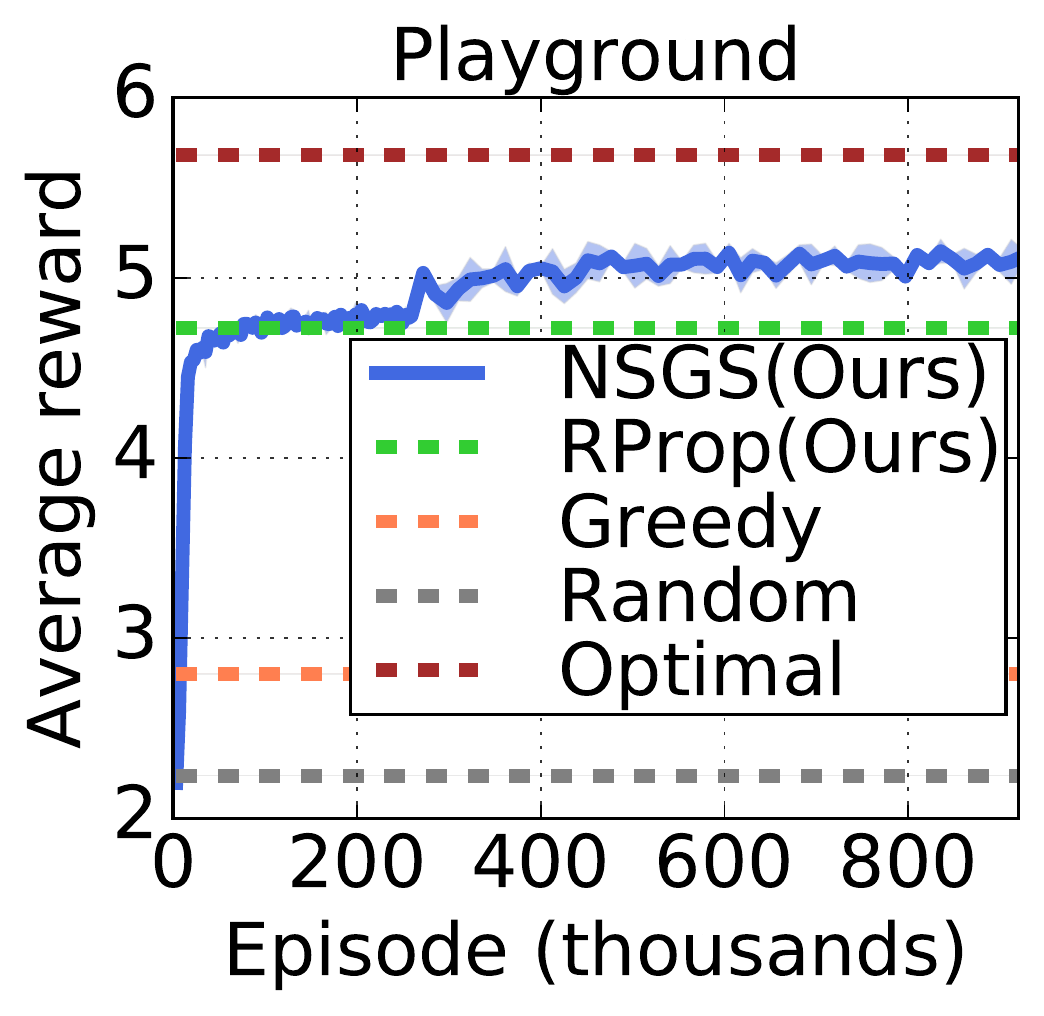}
  \vspace{-15pt}
  \caption{Learning curves on Mining and Playground domain. \NSGS{} is distilled from \RProp{} on 77K and 256K episodes, respectively, and finetuned after that.}
  \label{fig:learning-curve}
\vskip -0.1in
\end{wrapfigure}
\paragraph{Training Performance}
The learning curves of \NSGS{} and performance of other agents are shown in Figure~\ref{fig:learning-curve}. 
Our \RProp{} policy significantly outperforms the Greedy policy. 
This implies that the proposed idea of back-propagating the reward gradient captures long-term dependencies among subtasks to some extent. 
We also found that \NSGS{} further improves the performance through fine-tuning with actor-critic method. We hypothesize that \NSGS{} learned to estimate the expected costs of executing subtasks from the observations and consider them along with subtask graphs. 

 \begin{figure*}[p]
    \centering
   \includegraphics[width=0.95\linewidth]{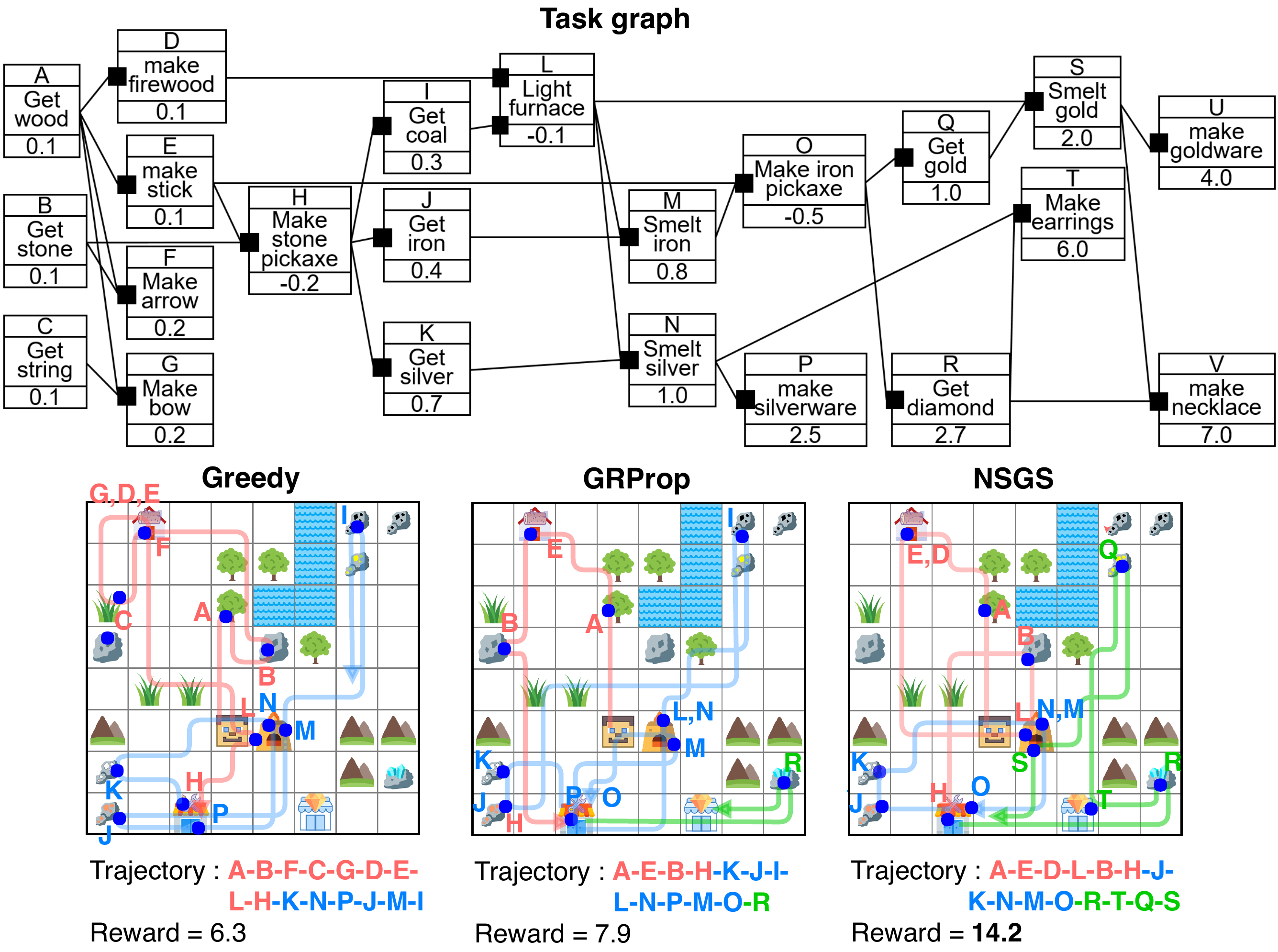}
   \vspace{-5pt}
\caption{Example trajectories of Greedy, \RProp{}, and \NSGS{} agents given 75 steps on Mining domain. We used different colors to indicate that agent has different types of pickaxes: red (no pickaxe), blue (stone pickaxe), and green (iron pickaxe). 
   Greedy agent prefers subtasks C, D, F, and G to H and L since C, D, F, and G gives positive immediate reward, whereas \NSGS{} and \RProp{} agents find a short path to make stone pickaxe, focusing on subtasks with higher long-term reward. Compared to \RProp{}, the \NSGS{} agent can find a shorter path to make an iron pickaxe, and succeeds to execute more number of subtasks.}
   \label{fig:qual}
   \vspace{-10pt}
\end{figure*}
\begin{figure*}[p]
    \centering
   \includegraphics[width=0.98\linewidth]{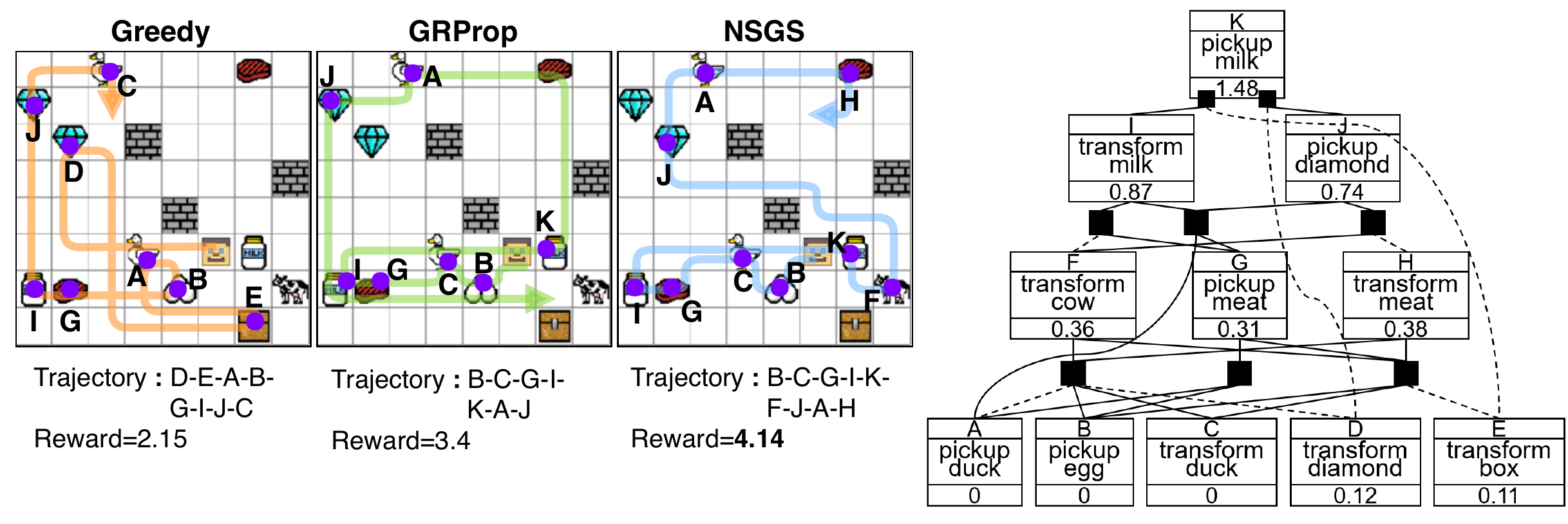}
   \vspace{-5pt}
   \caption{Example trajectories of Greedy, \RProp{}, and \NSGS{} agents given 45 steps on Playground domain. The subtask graph includes \texttt{NOT} operation and distractor (subtask D, E, and H). We removed stochasticity in environment for the controlled experiment. Greedy agent executes the distractors since they give positive immediate rewards, which makes it impossible to execute the subtask K which gives the largest reward. \RProp{} and \NSGS{} agents avoid distractors and successfully execute subtask K by satisfying its preconditions. After executing subtask K, the \NSGS{} agent found a shorter path to execute remaining subtasks than the \RProp{} agent and gets larger reward.}
   \label{fig:qual2}
   \vspace{-10pt}
\end{figure*}
 \begin{figure*}[!p]
  \centering
  \includegraphics[width=0.33\linewidth]{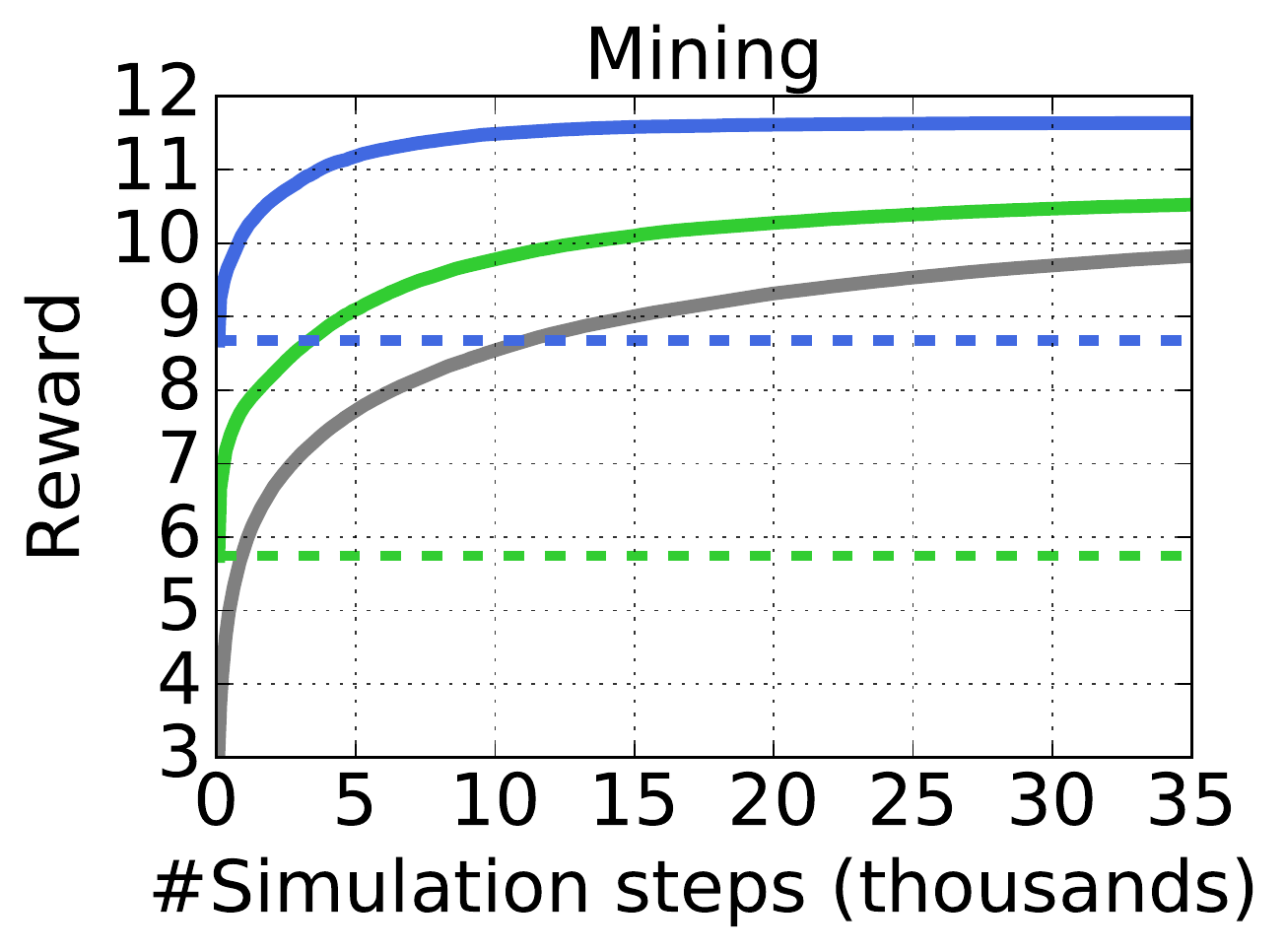}
  $\qquad$
  \includegraphics[width=0.33\linewidth]{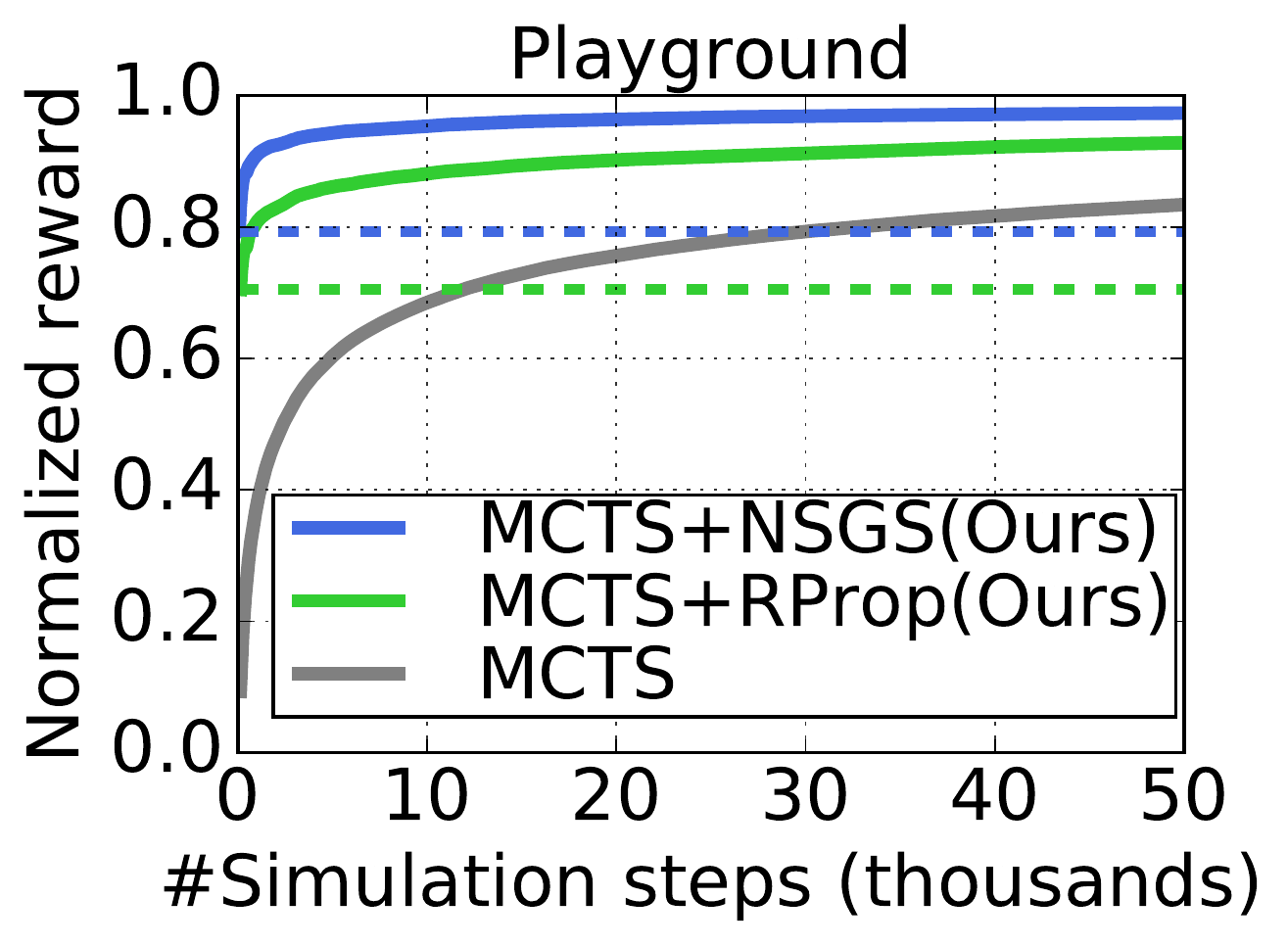}
  \vspace{-7pt}
  \caption{Performance of MCTS+\NSGS{}, MCTS+\RProp{} and MCTS per the number of simulated steps on (Left) \tb{Eval} of Mining domain and (Right) \tb{D2} of Playground domain (see Table~\ref{tab:depth}).}\label{fig:search}
\vspace{-10pt}
\end{figure*}
\cutparagraphup
\paragraph{Generalization Performance}
We considered two different types of generalization: a \tb{zero-shot} setting where agent must immediately achieve good performance on unseen subtask graphs without learning, and an \tb{adaptation} setting where agent can learn about task through the interaction with environment. Note that Independent agent was evaluated in adaptation setting only since it has no ability to generalize as it does not take subtask graph as input. Particularly, we tested agents on larger subtask graphs by varying the number of layers of the subtask graphs from four to six with a larger number of subtasks on Playground domain. Table~\ref{tab:depth} summarizes the results in terms of normalized reward $\bar{R}=(R-R_{min})/(R_{max}-R_{min})$ where $R_{min}$ and $R_{max}$ correspond to the average reward of the Random and the Optimal policy respectively. Due to large number of subtasks ($>$16) in Mining domain, the Optimal policy was intractable to be evaluated. Instead, we reported the un-normalized mean reward. Though the performance degrades as the subtask graph becomes larger as expected, \NSGS{} generalizes well to larger subtask graphs and consistently outperforms all the other agents on Playground and Mining domains in zero-shot setting. 
In adaptation setting, \NSGS{} performs slightly better than zero-shot setting by fine-tuning on the subtask graphs in evaluation set. Independent agent learned a policy comparable to Greedy, but performs much worse than \NSGS{}.

\cutsubsectionup
\subsection{ Qualitative Result }
\cutsubsectiondown
Figure~\ref{fig:qual} visualizes trajectories of agents on Mining domain. Greedy policy mostly focuses on subtasks with immediate rewards (e.g., get string, make bow) that are sub-optimal in the long run. In contrast, \NSGS{} and \RProp{} agents focus on executing subtask H (make stone pickaxe) in order to collect materials much faster in the long run. Compared to \RProp{}, \NSGS{} learns to consider observation also and avoids subtasks with high cost (e.g., get coal).\\
Figure~\ref{fig:qual2} visualizes trajectories on Playground domain. In this graph, there are distractors (e.g., D, E, and H) and the reward is delayed. In the beginning, Greedy chooses to execute distractors, since they gives positive reward while subtasks A, B, and C do not. However, \RProp{} observes non-zero gradient for subtasks A, B, and C that are propagated from the parent nodes. Thus, even though the reward is delayed, \RProp{} can figure out which subtask to execute. \NSGS{} learns to understand long-term dependencies from \RProp{}, and finds shorter path by also considering the observation.
\cutsubsectionup
\subsection{Combining \NSGS{} with Monte-Carlo Tree Search}
\label{sec:MCTS}
\cutsubsectiondown
We further investigated how well our \NSGS{} agent performs compared to conventional search-based methods and how our \NSGS{} agent can be combined with search-based methods to further improve the performance. We implemented the following methods (see Appendix for the detail):
\vspace{-6pt}
\begin{itemize}[leftmargin=*]
\setlength\itemsep{-0.1em}
    \item MCTS: An MCTS algorithm with UCB~\citep{auer2002finite} criterion for choosing actions.
    \item MCTS+\NSGS{}: An MCTS algorithm combined with our \NSGS{} agent. \NSGS{} policy was used as a rollout policy to explore reasonably good states during tree search, which is similar to AlphaGo~\citep{silver2016mastering}.
    \item MCTS+\RProp{}: An MCTS algorithm combined with our \RProp{} agent similar to MCTS+\NSGS{}.
\end{itemize}
\vspace{-6pt}
The results are shown in Figure~\ref{fig:search}. It turns out that our \NSGS{} performs as well as MCTS method with approximately 32K simulations on Playground and 11K simulations on Mining domain, while \RProp{} performs as well as MCTS with approximately 11K simulations on Playground and 1K simulations on Mining domain.
This indicates that our \NSGS{} agent implicitly performs long-term reasoning that is not easily achievable by a sophisticated MCTS, even though \NSGS{} does not use any simulation and has never seen such subtask graphs during training.
More interestingly, MCTS+\NSGS{} and MCTS+\RProp{} significantly outperforms MCTS, and MCTS+\NSGS{} achieves approximately $0.97$ normalized reward with 33K simulations on Playground domain. We found that the Optimal policy, which corresponds to normalized reward of $1.0$, uses approximately 648M simulations on Playground domain. Thus, MCTS+\NSGS{} performs almost as well as the Optimal policy with only $0.005\%$ simulations compared to the Optimal policy. This result implies that \NSGS{} can also be used to improve simulation-based planning methods by effectively reducing the search space. 
\cutsectionup
\section{Conclusion}\label{sec:c}
\cutsectiondown
We introduced the subtask graph execution problem which is an effective and principled framework of describing complex tasks. To address the difficulty of dealing with complex subtask dependencies, we proposed a graph reward propagation policy derived from a differentiable form of subtask graph, which plays an important role in pre-training our neural subtask graph solver architecture. The empirical results showed that our agent can deal with long-term dependencies between subtasks and generalize well to unseen subtask graphs. In addition, we showed that our agent can be used to effectively reduce the search space of MCTS so that the agent can find a near-optimal solution with a small number of simulations. 
In this paper, we assumed that the subtask graph (e.g., subtask dependencies and rewards) is given to the agent. However, it will be very interesting future work to investigate how to extend to more challenging scenarios where the subtask graph is unknown (or partially known) and thus need to be estimated through experience. 

\clearpage
\appendix
\section*{Acknowledgments}
%
This work was supported mainly by the ICT R\&D program of MSIP/IITP (2016-0-00563: Research on Adaptive Machine Learning Technology Development for Intelligent Autonomous Digital Companion) and partially by DARPA Explainable AI (XAI) program \#313498 and Sloan Research Fellowship.

\bibliographystyle{unsrtnat}
\bibliography{submission}

\clearpage
\section{Details of the Task}
We define each task as an MDP tuple $\mc{M}_{G}=(\mc{S, A}, \mc{P}_{G}, \mc{R}_{G}, \rho_{G},\gamma)$ where $\mc{S}$ is a set of states, $\mc{A}$ is a set of actions, $\mc{P}_{G}: \mc{S\times A\times S}\rightarrow [0,1]$ is a task-specific state transition function, $\mc{R}_{G}: \mc{S\times A}\rightarrow \mbb{R}$ is a task-specific reward function and $\rho_{G}: \mc{S}\rightarrow [0,1]$ is a task-specific initial distribution over states. 
We describe the subtask graph $G$ and each component of MDP in the following paragraphs.
\cutparagraphup
\paragraph{Subtask and Subtask Graph}  The subtask graph consists of $N$ subtasks that is a subset of $\mathcal{O}$, the subtask reward $\mb{r}\in\mbb{R}^{N}$, and the precondition of each subtask. The set of subtasks is $\mathcal{O}=\mathcal{A}_{int} \times \mathcal{X}$, where $\mathcal{A}_{int}$ is a set of primitive actions to interact with objects, and $\mathcal{X}$ is a set of all types of interactive objects in the domain. To execute a subtask $(a_{int}, obj) \in \mathcal{A}_{int} \times \mathcal{X}$, the agent should move on to the target object $obj$ and take the primitive action $a_{int}$.
\cutparagraphup
\paragraph{State} The state $\mb{s}_{t}$ consists of the observation $\mb{obs}_t\in\{0,1\}^{W\times H\times C}$, the completion vector $\mb{x}_t\in\{0,1\}^{N}$, the time budget $step_t$ and the eligibility vector $\mb{e}_t\in\{0,1\}^{N}$. An observation $\mb{obs}_t$ is represented as $H\times W\times C$ tensor, where $H$ and $W$ are the height and width of map respectively, and $C$ is the number of object types in the domain. The $(h,w,c)$-th element of observation tensor is $1$ if there is an object $c$ in $(h,w)$ on the map, and $0$ otherwise. The time budget indicates the number of remaining time-steps until the episode termination. The completion vector and eligibility vector provides additional information about $N$ subtasks. The details of completion vector and eligibility vector will be explained in the following paragraph.
\cutparagraphup
\paragraph{State Distribution and Transition Function} Given the current state $(\mb{obs}_t, \mb{x}_t, \mb{e}_t)$, the next step state $(\mb{obs}_{t+1}, \mb{x}_{t+1}, \mb{e}_{t+1})$ is computed from the subtask graph $G$. In the beginning of episode, the initial time budget $step_t$ is sampled from a pre-specified range $N_{step}$ for each subtask graph (See section~\ref{sec:task_graph_gen} for detail), the completion vector $\mb{x}_t$ is initialized to a zero vector in the beginning of the episode $\mb{x}_0=[0,\ldots,0]$ and the observation $\mb{obs}_0$ is sampled from the task-specific initial state distribution $\rho_{G}$. Specifically, the observation is generated by randomly placing the agent and the $N$ objects corresponding to the $N$ subtasks defined in the subtask graph $G$. When the agent executes subtask $i$, the $i$-th element of completion vector is updated by the following update rule:
\begin{align}
    x^i_{t+1} &=\left\{ \begin{array}{rcl}
1 & \mbox{if} & e_t^i=1 \\ 
x^i_t & \mbox{otherwise} & 
\end{array}\right..
\end{align}
The observation is updated such that agent moves on to the target object, and perform corresnponding primitive action (See Section~\ref{sec:env} for the full list of subtasks and corresponding primitive actions on Mining and Playground domain). The eligibility vector $\mb{e}_{t+1}$ is computed from the completion vector $\mb{x}_{t+1}$ and subtask graph $G$ as follows:
\begin{align}
  e_{t+1}^{i} &= \underset{j\in Child_i}{\text{OR}} \left( y^{j}_{AND}\right),\\
  y^{i}_{AND} &= \underset{j\in Child_i}{\text{AND}} \left( \widehat{x}_{t+1}^{i,j}\right),\\
  \widehat{x}_{t+1}^{i,j}&= x_{t+1}^jw^{i,j} + (1-x_{t+1}^j)(1-w^{i,j}),
  \label{eq:xhat-sup}
\end{align}
where $w^{i, j}=0$ if there is a \texttt{NOT} connection between $i$-th node and $j$-th node, otherwise $w^{i,j}=1$. Intuitively, $\widehat{x}_{t}^{i,j}=1$ when $j$-th node does not violate the precondition of $i$-th node. Executing each subtask costs different amount of time depending on the map configuration. Specifically, the time cost is given as the Manhattan distance between agent location and target object location in the grid-world plus one more step for performing a primitive action. 
\cutparagraphup
\paragraph{Task-specific Reward Function} The reward function is defined in terms of the subtask reward vector $\mb{r}$ and the eligibility vector $\mb{e}_t$, where the subtask reward vector $\mb{r}$ is the component of subtask graph $G$ the and eligibility vector is computed from the completion vector $\mb{x}_t$ and subtask graph $G$ as Eq.~\ref{eq:xhat-sup}. Specifically, when agent executes subtask $i$, the reward given to agent at time step $t$ is given as follows:
\begin{align}
    r_{t} &=\left\{ \begin{array}{rcl}
r^i & \mbox{if} & e_t^i=1 \\ 
0 & \mbox{otherwise} & 
\end{array}\right..
\end{align}

\section{Experiment on Hierarchical Task Network}
We compared with our methods with the recent graph-based multitask RL works~\cite{hayes2016autonomously, ghazanfari2017autonomous, huang2018neural}. However, these methods cannot be applied to our problem for two main reasons: 1) they aim to solve a single-goal task, which means they can only solve a subset of our problem, and 2) they require search or learning during test time, which means they cannot be applied in zero-shot generalization setting. Specifically, each trajectory in single-goal task is assumed to be labeled as success or failure depending on whether the goal was achieved or not, which is necessary for these methods~\cite{hayes2016autonomously, ghazanfari2017autonomous, huang2018neural} to infer the task structure (e.g., hierarchical task network (HTN)~\citep{sacerdoti1975nonlinear}). 
Since our task setting is more general and not limited to a single goal task, the task structure with multiple goals cannot be inferred with these methods.

For a direct comparison, we simplified our problem into single-goal task as follows. 1) We set a single goal; set all the subtask reward to 0, except the top-level subtask, and set it as terminal state. 2) We removed the cost, time budget, and observation, and set $\gamma=1$. After constructing the task network such as HTN, these methods~\cite{hayes2016autonomously, ghazanfari2017autonomous, huang2018neural} execute task by planning~\cite{hayes2016autonomously} or learning a policy~\cite{ghazanfari2017autonomous, huang2018neural} during test stage.
Accordingly, we evaluated HTN-plan method~\cite{hayes2016autonomously} in planning setting, and allowed learning in test time for~\cite{ghazanfari2017autonomous, huang2018neural}. 
Note that these methods cannot execute a task in zero-shot setting, while our NSGS can do it by learning an embedding of subtask graph; it is the main reason why our method performs much better than these methods in the following two experiments.

\noindent\makebox[\textwidth][c]{
\begin{minipage}{0.7\textwidth}
  \begin{minipage}[b]{0.49\textwidth}
    \centering
    \includegraphics[width=\linewidth]{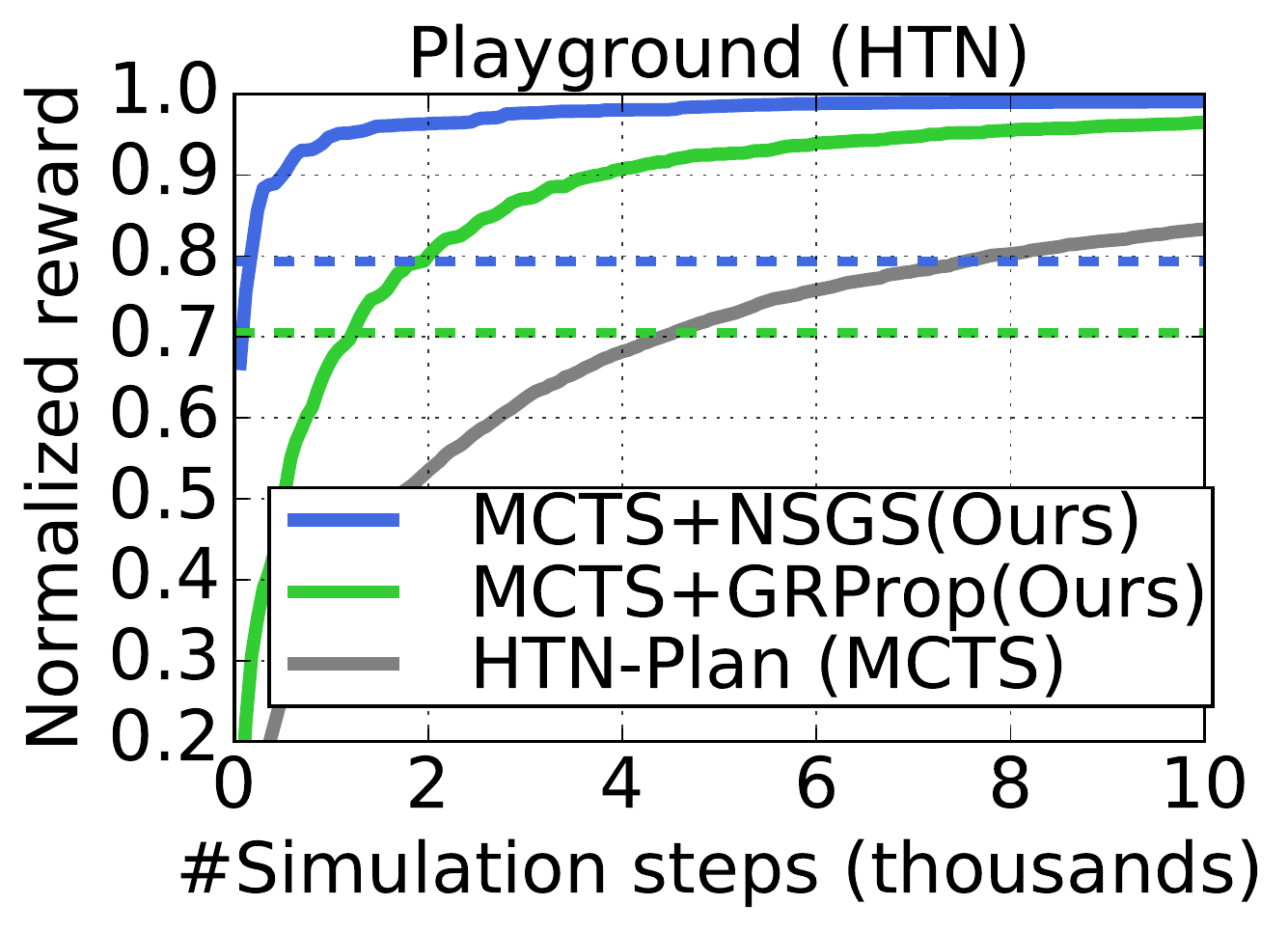}
    \captionof{figure}{Planning performance of MCTS+NSGS, MCTS+\RProp{} and HTN-Plan on HTN subtask graph in Playground domain.}
  \end{minipage}
  \hfill
  \begin{minipage}[b]{0.49\textwidth}
    \centering
    \begin{tabular}{|c|c|}
        \hline
        \multicolumn{2}{|c|}{Adaptation (HTN)}\\ \hlineB{2}
        Method  & $\bar{R}$   \\ \hline
        NSGS (Ours)  &\tb{.90}   \\
        HTN-Independent& .31    \\
      \hline
      \end{tabular}
      \captionof{table}{Adaptation performance (normalized reward) of NSGS and HTN-Independent on HTN subtask graph in Playground domain.}
    \end{minipage}
  \end{minipage}
}
\subsection{Comparison with HTN-Planning}
\citet{hayes2016autonomously} performed planning on the inferred task network to find the optimal solution. Thus, we implemented \tb{HTN-Plan} with MCTS as in section 5.5, and compared with ours in planning setting. We evaluated our \tb{MCTS+NSGS} and \tb{MCTS+\RProp{}} for comparison. The figure shows that our \tb{MCTS+NSGS} and \tb{MCTS+\RProp{}} agents outperform HTN-Plan by a large margin.

\subsection{Comparison with HTN-based Agent}
Instead of planning, \citet{ghazanfari2017autonomous} learned an hierarchical RL (HRL) agent on the constructed HTN during testing. Thus, we evaluated it in adaptation setting (i.e., learning during test time). To this end, we implemented an HRL agent, HTN-Independent, which is a policy over option trained on each subtask graph independently, similar to Independent agent (see section 5.2). 
The result shows that our NSGS agent can find the solution much faster than HTN-Independent agent due to zero-shot generalization ability. 

\citet{huang2018neural} inferred the subtask graph from the visual demonstration in testing. Since the environment state is available in our setting, providing demonstration amounts to providing the solution. Thus we couldn't compare with it.

\section{Details of \NSGS{} Architecture}\label{sec:NSS-detail}
\begin{figure}[thbp]
\vspace{-6pt}
    \centering
   \includegraphics[width=0.8\linewidth]{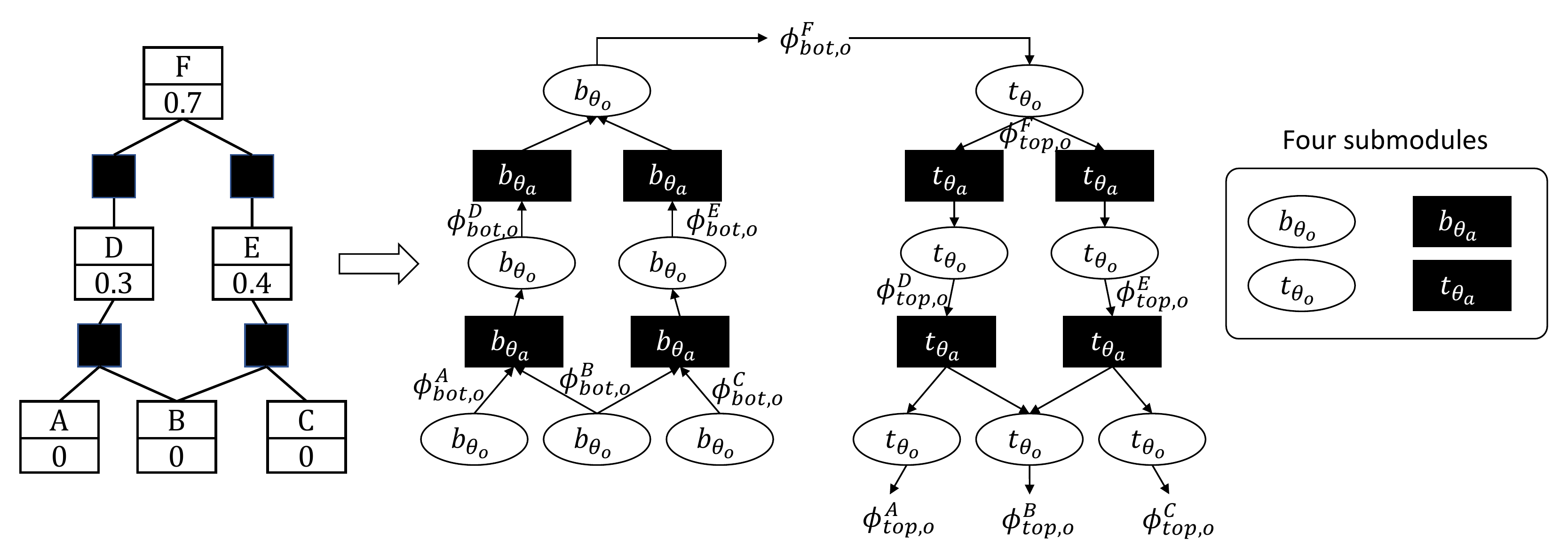}
   \caption{An example of R3NN construction for a given subtask graph input. The four encoders ($b_{\theta_a}, b_{\theta_o}, t_{\theta_a},$ and $t_{\theta_o}$) are cloned and connected according to the input subtask graph where the cloned models share the weight. For simplicity, only the output embeddings of bottom-up and top-down \texttt{OR} encoder were specified in the figure.
   }
   \label{fig:R3NN}
\end{figure}

\paragraph{Task module} Figure~\ref{fig:R3NN} illustrates the structure of the task module of \NSGS{} architecture for a given input subtask graph. Specifically, the task module was implemented with four encoders: $b_{\theta_a}, b_{\theta_o}, t_{\theta_a},$ and $t_{\theta_o}$. The input and output of each encoder is defined in the main text section \tb{4.1}  as:
\begin{align}
 \phi^{i}_{bot,o} & = b_{\theta_o}\left(x^{i}_t, e^{i}_t, step, \sum_{j \in Child_i}\phi^{j}_{bot,a} \right),&
 \phi^{j}_{bot,a} & = b_{\theta_a}\left(\sum_{k \in Child_j}\left[\phi^{k}_{bot,o},w_{+}^{j,k}\right]\right),\label{eq:rcnn22}\\
 \phi^{i}_{top,o} & = t_{\theta_o}\left(\phi^{i}_{bot,o}, r^i, \sum_{j \in Parent_i}\left[\phi^{j}_{top,a},w_{+}^{i,j}\right]\right), &
 \phi^{j}_{top,a} & = t_{\theta_a}\left( \phi^{j}_{bot,a}, \sum_{k \in Parent_j}\phi^{k}_{top,o}\right),\label{eq:rcnn44}
\end{align}
For bottom-up process, the encoder takes the output embeddings of its children encoders as input. Similarly, for top-down process, the encoder takes the output embeddings of its parent encoders as input. The input embeddings are aggregated by taking element-wise summation. For $\phi^{i}_{bot,a}$ and $\phi^{i}_{top,o}$, the embeddings are concatenated with $w_{+}^{i,j}$ to deal with \texttt{NOT} connection before taking the element-wise summation. Then, the summed embedding is concatenated with all additional input as defined in Eq.~\ref{eq:rcnn22} and~\ref{eq:rcnn44}, which is further transformed with three fully-connected layers with 128 units. The last fully-connected layer outputs 128-dimensional output embedding.
The embeddings are transformed to reward scores as via: $\mb{p}^{reward}_t = \boldsymbol{\Phi}_{top}^{\top}\mb{v},$
where $\boldsymbol{\Phi}_{top} = [\phi_{top,o}^1,\ldots,\phi_{top,o}^N]\in \mathbb{R}^{E \times N}$, $E$ is the dimension of the top-down embedding of \texttt{OR} node, and $\mb{v}\in \mathbb{R}^{E}$ is a weight vector for reward scoring. Similarly, the reward baseline is computed by $b^{reward}_t = \textrm{sum}(\boldsymbol{\Phi}_{top}^{\top}\tilde{\mb{v}})$, where sum($\cdot$) is the reduced-sum operation and $\tilde{\mb{v}}$ is the weight vector for reward baseline. We used parametric ReLU (PReLU) function as activation function.
\paragraph{Observation module}
 The network consists of BN1-Conv1(16x1x1-1/0)-BN2-Conv2(32x3x3-1/1)-BN3-Conv3(64x3x3-1/1)-BN4-Conv4(96x3x3-1/1)-BN5-Conv5(128x3x3-1/1)-BN6-Conv6(64x1x1-1/0)-FC(256). The output embedding of FC(256) was then concatenated with the number of remaining time step $step_t$. Finally, the network has two fully-connected output layers for the cost score $\mb{p}_{t}^{cost}\in\mbb{R}^{N}$ and the cost baseline $b_{t}^{cost}\in\mbb{R}$. Then, the policy of \NSGS{} is calculated by adding reward score and cost score, and taking softmax:
 \begin{align}
     \pi(\mb{o}_{t}|\mb{s}_{t}, G) = \text{Softmax}( \mb{p}^{reward}_t+\mb{p}^{cost}_t ).
 \end{align}
 The baseline output is obtained by adding reward baseline and cost baseline:
 \begin{align}
     V_{\theta'}(\mb{s}_{t}, G) = b^{reward}_t+b^{cost}_t.
 \end{align}

\section{Details of Learning \NSGS{} Agent}
\paragraph{Learning objectives}
The \NSGS{} architecture is first trained through policy distillation and finetuned using actor-critic method with generalized advantage estimator.
During policy distillation, the KL divergence between \NSGS{} and teacher policy (\RProp{}) is minimized as follows:
\begin{align}
    \nabla_{\theta} \mathcal{L}_{1} = \mathbb{E}_{G \sim \mathcal{G}_{train}}\left[\mathbb{E}_{s \sim \pi^{G}_{\theta}}\left[\nabla_{\theta} D_{KL}\left(\pi_T^{G} || \pi^{G}_{\theta} \right)  \right] \right],
\end{align}
where $\theta$ is the parameter of \NSGS{} architecture, $\pi^{G}_{\theta}$ is the simplified notation of \NSGS{} policy with subtask graph input $G$, $\pi_T^{G}$ is the simplified notation of teacher (\RProp{}) policy with subtask graph input $G$, $D_{KL}\left(\pi_T^{G} || \pi^{G}_{\theta}\right) = \sum_{a} \pi_T^{G} \log \frac{\pi_T^{G}}{\pi^{G}_{\theta}}$ and $\mathcal{G}_{train}\subset \mathcal{G}$ is the training set of subtask graphs.

For both policy distillation and fine-tuning, we sampled one subtask graph for each 16 parallel workers, and each worker in turn sample a mini-batch of 16 world configurations (maps). Then, \NSGS{} generates total 256 episodes in parallel. After generating episode, the gradient from 256 episodes are collected and averaged, and then back-propagated to update the parameter. For policy distillation, we trained \NSGS{} for 40 epochs where each epoch involves 100 times of update. Since our \RProp{} policy observes only the subtask graph, we only trained task module during policy distillation. The observation module was trained for auxiliary prediction task; observation module predicts the number of step taken by agent to execute each subtask.

 After policy distillation, we finetune \NSGS{} agent in an end-to-end manner using actor-critic method with generalized advantage estimation (GAE)~\citep{schulman2015high} as follows:
\begin{align}
\nabla_{\theta} \mathcal{L}_{2} &= \mathbb{E}_{G\sim \mathcal{G}_{train}}\left[\mathbb{E}_{s \sim \pi^{G}_{\theta}} \left[ -\nabla_{\theta}\log\pi^{G}_{\theta}\sum^{\infty}_{l=0} \left(\prod_{n=0}^{l-1}{(\gamma\lambda)^{k_n}}\right) \delta_{t+l} \right] \right],\\
\delta_t &= r_t + \gamma^{k_t} V^{\pi}_{\theta'}(\mb{s}_{t+1},G)  - V^{\pi}_{\theta'}(\mb{s}_{t},G),
\end{align} 
where $k_t$ is the duration of option $\mb{o}_t$, $\gamma$ is a discount factor, $\lambda \in \left[0, 1\right]$ is a weight for balancing between bias and variance of the advantage estimation, and $V_{\theta'}^{\pi}$ is the critic network parameterized by $\theta'$. 
During training, we update the critic network to minimize $\mathbb{E}\left[ \left(R_t - V^{\pi}_{\theta'}(\mb{s}_{t},G) \right)^2 \right]$, where $R_t$ is the discounted cumulative reward at time $t$.
\paragraph{Hyperparameters}
For both finetuning and policy distillation, we used RMSProp optimizer with the smoothing parameter of 0.97 and epsilon of 1e-6. When distilling agent with teacher policy, we used learning rate=1e-4 and multiplied it by 0.97 on every epoch for both Mining and Playground domain. For finetuning, we used learning rate=2.5e-6 for Playground domain, and 2e-7 for Mining domain. For actor-critic training for \NSGS{}, we used $\alpha=0.03,\ \lambda=0.96,\ \gamma=0.99$.

\section{Details of AND/OR Operation and Approximated AND/OR Operation}
{
}
 In section 4.2, the output of $i$-th AND and OR node in subtask graph were defined using AND and OR operation with multiple input. They can be represented in logical expression as below:
 \begin{align}
 \underset{j\in Child_i}{\text{OR}} \left( y^{j}\right) &=y^{j_{1}}\vee y^{j_{2}}\vee \ldots\vee y^{j_{|Child_i|}},\\
 \underset{j\in Child_i}{\text{AND}} \left( y^{j}\right)&=y^{j_{1}}\wedge y^{j_{2}}\wedge \ldots\wedge y^{j_{|Child_i|}},
 \end{align}
 where $j_1,\ldots,j_{|Child_i|}$ are the elements of a set $Child_i$ and $Child_i$ is the set of inputs coming from the children nodes of $i$-th node. Then, these AND and OR operations are smoothed as below:
 \begin{align}
 \underset{j\in Child_i}{\wt{\text{OR}}} \left( \wt{y}^{j}_{AND}\right)
 &=h_{or}\left( \sum_{j\in Child_i}{ \wt{y}^{j}_{AND} } \right),\\
 \underset{j\in Child_i}{\wt{\text{AND}}} \left( \widehat{x}_{t}^{i,j}\right) &=h_{and} \left( \sum_{j\in Child_i}{ \widehat{x}_{t}^{i,j} } - |Child_i| + 0.5 \right), \label{eq:andor3}
 \end{align}
 where $h_{or}(x) = \alpha_{o} \tr{tanh}(x/\beta_{o})$, $h_{and}(x) = \alpha_{a} \sigma(x/\beta_{a})$, $\sigma(\cdot)$ is sigmoid function, and $\alpha_o, \beta_o, \alpha_a, \beta_a\in \mbb{R}$ are hyperparameters to be set. We used $\beta_a=0.6, \beta_o=2, \alpha_a=1/\sigma(0.25), \alpha_o=1$ for Mining domain, and $\beta_a=0.5, \beta_o=1.5, \alpha_a=1/\sigma(0.25), \alpha_o=1$ for Playground domain.
\section{Details of Subtask Executor}\label{sec:SE}
\paragraph{Architecture}
The subtask executor has the same architecture of the parameterized skill architecture of~\cite{oh2017zero} with slightly different hyperparameters. The network consists of Conv1(32x3x3-1/1)-Conv2(32x3x3-1/1)-Conv3(32x1x1-1/0)-Conv4(32x3x3-1/1)-LSTM(256)-FC(256). The subtask executor takes two task parameters $(q =[q^{(1)}, q^{(2)}])$ as additional input and computes $\chi(q) = \text{ReLU}(W^{(1)}q^{(1)} \odot W^{(2)}q^{(2)})$ to compute the subtask embedding, and further linearly transformed into the weights of Conv3 and the (factorized) weight of LSTM through multiplicative interaction as described above. Finally, the network has three fully-connected output layers for actions, termination probability, and baseline, respectively.
\paragraph{Learning objective}
The subtask executor is trained through policy distillation and then finetuned. Similar to~\citep{oh2017zero}, we first trained 16 teacher policy network for each subtask. The teacher policy network consists of Conv1(16x3x3-1/1)-BN1(16)-Conv2(16x3x3-1/1)-BN2(16)-Conv3(16x3x3-1/1)-BN3(16)-LSTM(128)-FC(128). Similar to subtask executor network, the teacher policy network has three fully-connected output layers for actions, termination probability, and baseline, respectively.
Then, the learned teacher policy networks are used as teacher policy for policy distillation to train subtask executor. During policy distillation, we train agent to minimize the following objective function:
\begin{align}
    \nabla_{\xi} \mathcal{L}_{1,sub} = \mathbb{E}_{\mb{o} \sim \mathcal{O}}\left[\mathbb{E}_{s \sim \pi^{\mb{o}}_{\xi}}\left[\nabla_{\xi}\left\{ D_{KL}\left(\pi_T^{\mb{o}} || \pi^{\mb{o}}_{\xi} \right)  \right]+\alpha L_{term}\right\} \right],
\end{align}
where $\xi$ is the parameter of subtask executor network, $\pi^{\mb{o}}_{\xi}$ is the simplified notation of subtask executor given input subtask $\mb{o}$, $\pi_T^{\mb{o}}$ is the simplified notation of teacher policy for subtask $\mb{o}$, $L_{term}=-\mbb{E}_{\mb{s}_t \in \tau_{\mb{o}}}\left[\log\beta_{\xi}(\mb{s}_{t},\mb{o})\right]$ is the cross entropy loss of predicting termination, $\tau_{\mb{o}}$ is a set of state in which the subtask $\mb{o}$ is terminated, $\beta_{\xi}(s_{t},\mb{o})$ is the termination probability output, and $D_{KL}\left(\pi_T^{\mb{o}} || \pi^{\mb{o}}_{\xi}\right) = \sum_{a} \pi_T^{\mb{o}}(a|s) \log \frac{\pi_T^{\mb{o}}(a|s)}{\pi^{\mb{o}}_{\xi} (a|s)}$.
After policy distillation, we finetuned subtask executor using actor-critic method with generalized advantage estimation (GAE):
\begin{equation}
\nabla_{\xi} \mathcal{L}_{2,sub} = 
\mathbb{E}_{\mb{o}\sim \mathcal{O}}\left[\mathbb{E}_{s \sim \pi^{\mb{o}}_{\xi}} \left[ -\nabla_{\xi}\log\pi_{\xi}\left(\mb{a}_{t}|\mb{obs}_{t},\mb{o}\right)\sum^{\infty}_{k=0}(\gamma \lambda)^k \delta_{t+k}+\alpha\nabla_{\xi}L_{term} \right] \right],
\end{equation}
where $\gamma \in \left[0, 1\right]$ is a discount factor, $\lambda \in \left[0, 1\right]$ is a weight for balancing between bias and variance of the advantage estimation, and $\delta_t=r_t + \gamma V^{\pi}(\mb{obs}_{t+1};\xi ')-V^{\pi}(\mb{obs}_{t}; \xi ')$. We used $\lambda=0.96,\ \gamma=0.99$ for fine-tuning, and $\alpha=0.1$ for both policy distillation and fine-tuning.
\section{Details of LSTM Baseline}
\paragraph{Architecture}
The LSTM baseline consists of LSTM on top of CNN. The architecture of CNN is the same as the CNN architecture of observation module of \NSGS{} described in the section~\ref{sec:NSS-detail}, and the architecture of LSTM is the same as the LSTM architecture used in subtask executor described in the section~\ref{sec:SE}. Specifically, it consists of BN1-Conv1(16x1x1-1/0)-BN2-Conv2(32x3x3-1/1)-BN3-Conv3(64x3x3-1/1)-BN4-Conv4(96x3x3-1/1)-BN5-Conv5(128x3x3-1/1)-BN6-Conv6(64x1x1-1/0)-LSTM(256)-FC(256).
The CNN takes the observation tensor as an input and outputs an embedding. The embedding is then concatenated with other input vectors including subtask completion indicator $\mb{x}_t$, eligibility vector $\mb{e}_t$, and the remaining step $step_t$. Finally, LSTM takes the concatenated vector as an input and output the softmax policy with the parameter $\theta^{'}$: $\pi_{\theta^{'}}\left(\mb{o}_{t}|\mb{obs}_{t},\mb{x}_t,\mb{e}_t,step_t\right)$.

\paragraph{Learning objective}
The LSTM baseline was trained using actor-critic method. For the baseline, we found that the moving average of return works much better than learning a critic network, and used it for experiment. This is due to the characteristic of adaptation setting; in adaptation setting, the subtask graph is fixed and the agent is trained for only a small number of episodes such that the critic network is usually under-fitted. Similar to \NSGS{}, the learning objective is given as
\begin{equation}
\nabla_{\theta^{'}} \mathcal{L}_{LSTM} = \mathbb{E}_{s \sim \pi^{G}_{\theta^{'}}} \left[ -\nabla_{\theta^{'}}\log\pi_{\theta^{'}}\left(\mb{o}_{t}|\mb{obs}_{t},\mb{x}_t,\mb{e}_t,step_{t}\right)
\sum^{\infty}_{l=0} \left(\prod_{n=0}^{l-1}{(\gamma\lambda)^{k_n}}\right) \delta_{t+l} \right],
\end{equation}
where $\gamma \in \left[0, 1\right]$ is a discount factor, $\lambda \in \left[0, 1\right]$ is a weight for balancing between bias and variance of the advantage estimation, $\delta_t=r_t + \gamma^{k_t} \overline{V}(t+1)-\overline{V}(t)$, and $\overline{V}(t)$ is the moving average of return at time step $t$. We used $\lambda=0.96$ and $\gamma=0.99$.
\section{Details of Search Algorithms}\label{sec:mcts}
Each iteration of Monte-Carlo tree search method consists of four stages: selection, expansion, rollout, and back-propagation. 
\begin{itemize}
    \item Selection: We used UCB criterion~\cite{auer2002finite}. Specifically, the option for which the score below has the highest value is chosen for selection:
\begin{align}
    \tr{score}=\frac{R_i}{n_i} + C_{UCB}\sqrt{\frac{\tr{ln} N}{n_i}},
\end{align}
where $R_i$ is the accumulated return at $i$-th node, $n_i$ is the number of visit of $i$-th node, $C_{UCB}$ is the exploration-exploitation balancing weight, and $N$ is the number of total iterations so far.
We found that $C_{UCB}=2\sqrt{2}$ gives the best result and used it for MCTS, MCTS+\RProp{} and MCTS+\NSGS{} methods.

    \item Expansion: MCTS randomly chooses the remaining eligible subtask, while the subtask is chosen by \NSGS{} policy for MCTS+\NSGS{} method and \RProp{} policy for MTS+\RProp{} method. More specifically, MCTS+\NSGS{} and MCTS+\RProp{} greedily chooses among the remaining subtasks based on \NSGS{} and \RProp{} policy, respectively. Due to the memory limit, the expansion of search tree was truncated at the depth of 7 for Playground and 10 for Mining domains, and performed rollout after the maximum depth.
    
    \item Rollout: MCTS randomly executes an eligible subtask, while MCTS+\NSGS{} and MCTS+\RProp{} execute the subtask with the highest probability given by \NSGS{} and \RProp{} policies, respectively.
    
    \item Back-propagation: Once the episode is terminated, the result is back-propagated; the accumulated return $R_i$ and the visit count $n_i$ are updated for the nodes in the tree that agent visited within the episode, and the number of total iteration is updated as $N\leftarrow N+1$.
\end{itemize}

\section{Details of Environment}\label{sec:env}
\subsection{Mining}
There are 15 types of objects: \textit{Mountain}, \textit{Water}, \textit{Work space}, \textit{Furnace}, \textit{Tree}, \textit{Stone}, \textit{Grass}, \textit{Pig}, \textit{Coal}, \textit{Iron}, \textit{Silver}, \textit{Gold}, \textit{Diamond}, \textit{Jeweler's shop}, and \textit{Lumber shop}. The agent can take 10 primitive actions: \textit{up}, \textit{down}, \textit{left}, \textit{right}, \textit{pickup}, \textit{use1}, \textit{use2}, \textit{use3}, \textit{use4}, \textit{use5} and agent cannot moves on to the \textit{Mountain} and \textit{Water} cell. \textit{Pickup} removes the object under the agent, and \textit{use}'s do not change the observation. There are 26 subtasks in the Mining domain:
\begin{itemize}
\item Get wood/stone/string/pork/coal/iron/silver/gold/diamond: The agent should go to \textit{Tree}/\textit{Stone}/\textit{Grass}/\textit{Pig}/\textit{Coal}/\textit{Iron}/\textit{Silver}/\textit{Gold}/\textit{Diamond} respectively, and take \textit{pickup} action.
\item Make firewood/stick/arrow/bow: The agent should go to \textit{Lumber shop} and take \textit{use1}/\textit{use2}/\textit{use3}/\textit{use4} action respectively.
\item Light furnace: The agent should go to \textit{Furnace} and take \textit{use1} action.
\item Smelt iron/silver/gold: The agent should go to \textit{Furnace} and take \textit{use2}/\textit{use3}/\textit{use4} action respectively.
\item Make stone-pickaxe/iron-pickaxe/silverware/goldware/bracelet: The agent should go to \textit{Work space} and take \textit{use1}/\textit{use2}/\textit{use3}/\textit{use4}/\textit{use5} action respectively.
\item Make earrings/ring/necklace: The agent should go to \textit{Jeweler's shop} and take \textit{use1}/\textit{use2}/\textit{use3} action respectively.
\end{itemize}
The icons used in Mining domain were downloaded from \texttt{www.icons8.com} and \texttt{www.flaticon.com}. The \textit{Diamond} and \textit{Furnace} icons were made by Freepik from \texttt{www.flaticon.com}.

\subsection{Playground}
There are 10 types of objects: \textit{Cow}, \textit{Milk}, \textit{Duck}, \textit{Egg}, \textit{Diamond}, \textit{Heart}, \textit{Box}, \textit{Meat}, \textit{Block}, and \textit{Ice}. The \textit{Cow} and \textit{Duck} move by 1 pixel in random direction with the probability of 0.1 and 0.2, respectively. The agent can take 6 primitive actions: \textit{up}, \textit{down}, \textit{left}, \textit{right}, \textit{pickup}, \textit{transform} and agent cannot moves on to the \textit{block} cell. \textit{Pickup} removes the object under the agent, and \textit{transform} changes the object under the agent to \textit{Ice}. The subtask graph was randomly generated without any hand-coded template (see Section~\ref{sec:task_graph_gen} for details). 

\section{Details of Subtask Graph Generation} \label{sec:task_graph_gen}
\subsection{Mining Domain}
 \begin{figure}[!htp]
  \vspace{-6pt}
    \centering
   \includegraphics[width=0.95\linewidth]{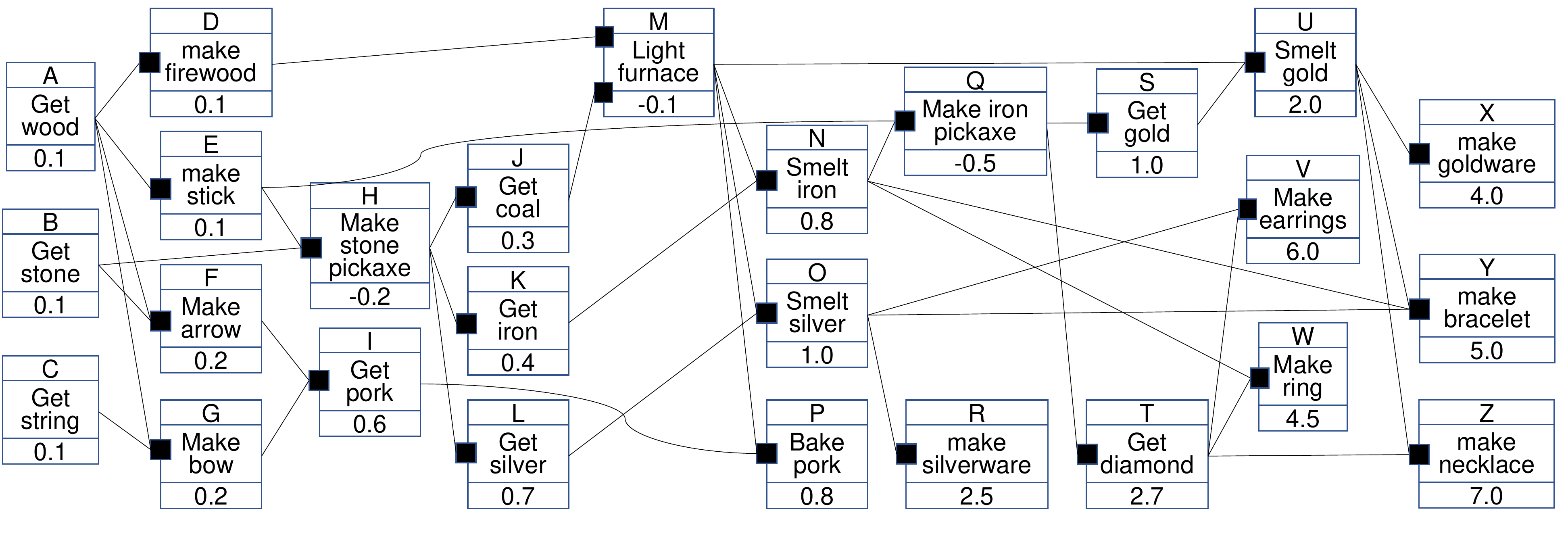}
   \caption{ The entire graph of Mining domain. Based on this graph, we generated 640 subtask graphs by removing the subtask node that has no parent node.}
   \label{fig:full_graph_mining}
\end{figure}
The precondition of each subtask in Mining domain was defined as Figure~\ref{fig:full_graph_mining}. Based on this graph, we generated all possible sub-graphs of it by removing the subtask node that has no parent node, while always keeping subtasks A, B, D, E, F, G, H, I, K, L. The reward of each subtask was randomly scaled by a factor of $0.8\sim 1.2$.
\subsection{Playground Domain}
\begin{table*}[!hbt] 
\centering
\begin{tabular}{r|c|l}
  \hline
      & $N_T$       & number of tasks in each layer \\ 
Nodes & $N_D$       & number of distractors in each layer \\
      & $N_A$       & number of \texttt{AND} node in each layer \\
      & $r$       & reward of subtasks in each layer \\
  \hline
      & $N_{ac}^{+}$& number of children of \texttt{AND} node in each layer \\ 
      & $N_{ac}^{-}$& number of children of \texttt{AND} node with \texttt{NOT} connection in each layer \\
Edges & $N_{dp}$      & number of parents with \texttt{NOT} connection of distractors in each layer \\
      & $N_{oc}$      & number of children of \texttt{OR} node in each layer \\
  \hline
Episode & $N_{step}$    & number of step given for each episode \\
    \hline
\end{tabular}
\vspace{-2pt}
\caption{Parameters for generating task including subtask graph parameter and episode length.}
\label{tb:graph_param_types}
\end{table*}
For training and test sample generation, the subtask graph structure was defined in terms of the parameters in table~\ref{tb:graph_param_types}. To cover wide range of subtask graphs, we randomly sampled the parameters $N_A, N_O, N_{ac}^{+}, N_{ac}^{-}, N_{dc}$, and $N_{oc}$ from the range specified in the table~\ref{tb:graph_param} and~\ref{tb:graph_param2}, while $N_T$ and $N_D$ was manually set. We prevented the graph from including the duplicated \texttt{AND} nodes with the same children node(s). We carefully set the range of each parameter such that at least 500 different subtask graphs can be generated with the given parameter ranges. The table~\ref{tb:graph_param} summarizes parameters used to generate training and evaluation subtask graphs for the Playground domain.

\begin{table*}[!hbt] 
\centering
\begin{tabular}{r|c|l}
  \hline
      & $N_T$       & \{6,4,2,1\}\\ 
      & $N_D$       & \{2,1,0,0\}\\
      & $N_A$       & \{3,3,2\}-\{5,4,2\}\\
Train & $N_{ac}^{+}$& \{1,1,1\}-\{3,3,3\}\\ 
(=\tb{D1})& $N_{ac}^{-}$& \{0,0,0\}-\{2,2,1\}\\
      & $N_{dp}$    &\{0,0,0\}-\{3,3,0\}\\
      & $N_{oc}$    &\{1,1,1\}-\{2,2,2\}\\
      & $r$       &\{0.1,0.3,0.7,1.8\}-\{0.2,0.4,0.9,2.0\}\\
      & $N_{step}$  &48-72\\
  \hline
      & $N_T$       & \{7,5,2,1\}\\ 
      & $N_D$       & \{2,2,0,0\}\\
      & $N_A$       & \{4,3,2\}-\{5,4,2\}\\
\tb{D2}& $N_{ac}^{+}$& \{1,1,1\}-\{3,3,3\}\\ 
      & $N_{ac}^{-}$& \{0,0,0\}-\{2,2,1\}\\
      & $N_{dp}$      &\{0,0,0,0\}-\{3,3,0,0\}\\
      & $N_{oc}$      &\{1,1,1\}-\{2,2,2\}\\
      & $r$       &\{0.1,0.3,0.7,1.8\}-\{0.2,0.4,0.9,2.0\}\\
      & $N_{step}$  &52-78\\
  \hline
      & $N_T$       & \{5,4,4,2,1\}\\ 
      & $N_D$       & \{1,1,1,0,0\}\\
      & $N_A$       & \{3,3,3,2\}-\{5,4,4,2\}\\
\tb{D3}& $N_{ac}^{+}$& \{1,1,1,1\}-\{3,3,3,3\}\\ 
      & $N_{ac}^{-}$& \{0,0,0,0\}-\{2,2,1,1\}\\
      & $N_{dp}$      &\{0,0,0,0,0\}-\{3,3,3,0,0\}\\
      & $N_{oc}$      &\{1,1,1,1\}-\{2,2,2,2\}\\
      & $r$       &\{0.1,0.3,0.6,1.0,2.0\}-\{0.2,0.4,0.7,1.2,2.2\}\\
      & $N_{step}$  &56-84\\
  \hline
      & $N_T$       & \{4,3,3,3,2,1\}\\ 
      & $N_D$       & \{0,0,0,0,0,0\}\\
      & $N_A$       & \{3,3,3,3,2\}-\{5,4,4,4,2\}\\
\tb{D4}& $N_{ac}^{+}$& \{1,1,1,1,1\}-\{3,3,3,3,3\}\\ 
      & $N_{ac}^{-}$& \{0,0,0,0,0\}-\{2,2,1,1,0\}\\
      & $N_{dp}$      &\{0,0,0,0,0,0\}-\{0,0,0,0,0,0\}\\
      & $N_{oc}$      &\{1,1,1,1,1\}-\{2,2,2,2,2\}\\
      & $r$       &\{0.1,0.3,0.6,1.0,1.4,2.4\}-\{0.2,0.4,0.7,1.2,1.6,2.6\}\\
      & $N_{step}$  &56-84\\
  \hline
  \end{tabular}
\vspace{-2pt}
\caption{Subtask graph parameters for training set and tasks \tb{D1}$\sim$\tb{D4}.}
\label{tb:graph_param}
\end{table*}

\section{Ablation Study on Neural Subtask Graph Solver Agent}\label{sec:ablation}
\subsection{Learning without Pre-training}
  \begin{table}[t] 
  \centering
  \small
      \begin{tabular}{|r|c|c|c|c|c|}
      \hlineB{2}
            \multicolumn{6}{|c|}{Zero-Shot Performance} \\\hlineB{2}
            &\multicolumn{4}{c|}{Playground($\bar{R}$)}    &Mining($R$)\\ \hline
         Task   &\tb{D1}    &\tb{D2}    &\tb{D3}    &\tb{D4}&\tb{Eval}\\ \hline
            
\NSGS{} (Ours)      &\tb{.820}  &\tb{.785} &\tb{.715} &\tb{.527}  &\tb{8.19}\\
\NSGS{}-task (Ours) & .773      & .730  & .645      &  .387     &6.51 \\
\RProp{} (Ours)    & .721      & .682  & .623      &  .424     &6.16 \\
\NSGS{}-scratch (Ours)& .046    & .056  & .062      &  .106     &3.68 \\
Random          & 0         & 0     & 0         &  0        &2.79 \\
      \hline
      \end{tabular}
      \caption{ Zero-shot generalization performance on Playground and Mining domain. \NSGS{}-scratch agent performs much worse than \NSGS{} and \RProp{} agent on Playground and Mining domain.}
  \label{tab:scratch}
  \end{table}
We implemented \tb{\NSGS{}-scratch} agent that is trained with actor-critic method from scratch without pre-training from \RProp{} policy to show that pre-training plays a crucial role for training our \NSGS{} agent. 
Table~\ref{tab:scratch} summarizes the result. \NSGS{}-scratch performs much worse than \NSGS{}, suggesting that pre-training is important in training \NSGS{}. This is not surprising as our problem is combinatorially intractable (e.g. searching over optimal sequence of subtasks given an unseen subtask graph).

\subsection{Ablation Study on the Balance between Task and Observation Module}
We implemented \tb{\NSGS{}-task} agent that uses only the task module without observation module to compare the contribution of task module and observation module of \NSGS{} agent. 
Overall, our \NSGS{} agent outperforms the \NSGS{}-task agent, showing that the observation module improves the performance by a large margin.
\section{Experiment Result on Subtask Graph Features}~\label{sec:features}
\begin{figure}[thbp]
\vspace{-6pt}
    \centering
   \includegraphics[width=0.8\linewidth]{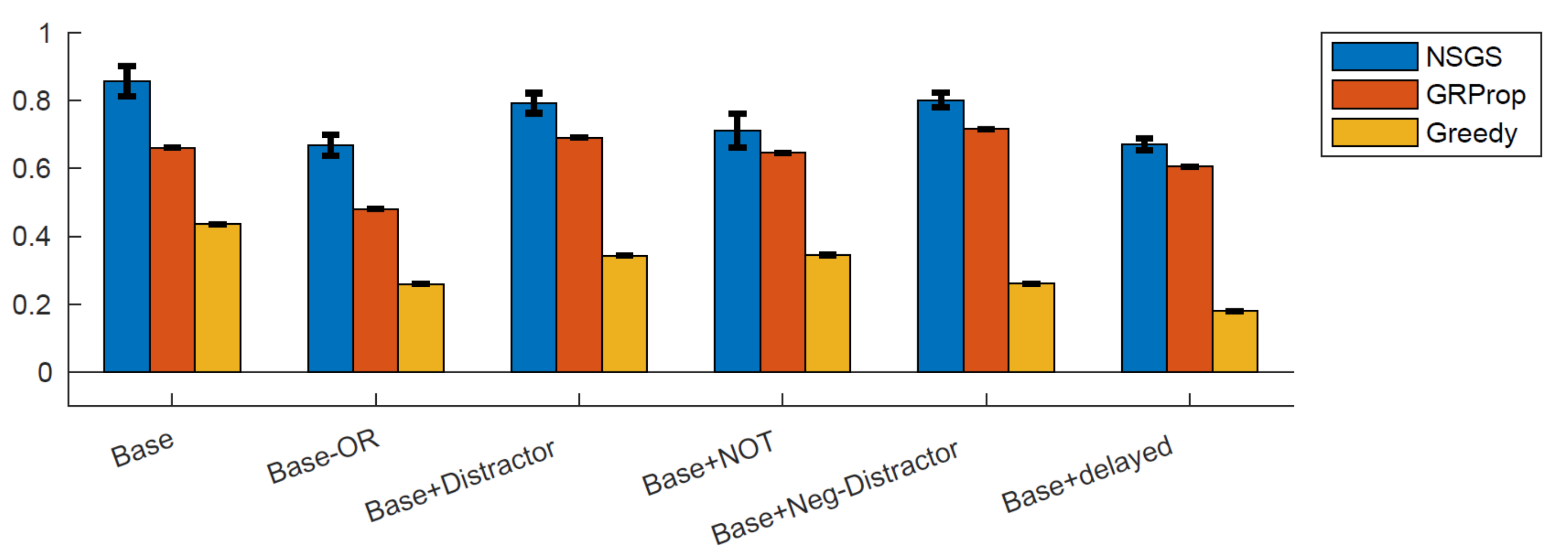}
   \caption{ Normalized performance on subtask graphs with different types of dependencies.}
   \label{fig:feature}
\end{figure}
To investigate how agents deal with different types of subtask graph components, we evaluated all agents on the following types of subtask graphs:
\begin{itemize}[leftmargin=*]
\item `Base' set consists of subtask graphs with \texttt{AND} and \texttt{OR} operations, but without \texttt{NOT} operation. 
\item `Base-OR' set removes all the \texttt{OR} operations from the base set.
\item `Base+Distractor' set adds several distractor subtasks to the base set.
\item `Base+NOT' set adds several \texttt{NOT} operations to the base set.
\item `Base+NegDistractor' set adds several negative distractor subtasks to the base set.
\item `Base+Delayed' set assigns zero reward to all subtasks but the top-layer subtask.
\end{itemize}
Note that we further divided the set of Distractor into Distractor and NegDistractor. The distractor subtask is a subtask without any parent node in the subtask graph. Executing this kind of subtask may give an immediate reward but is sub-optimal in the long run. The negative-distractor subtask is a subtask with only and at least one \texttt{NOT} connection to parent nodes in the subtask graph. Executing this subtask may give an immediate reward, but this would make other subtasks not executable.
Table~\ref{tb:graph_param2} summarizes the detailed parameters used for generating subtask graphs.
The results are shown in Figure~\ref{fig:feature}. Since `Base' and `Base-OR' sets do not contain \texttt{NOT} operation and every subtask gives a positive reward, the greedy baseline performs reasonably well compared to other sets of subtask graphs. It is also shown that the gap between \NSGS{} and \RProp{} is relatively large in these two sets. This is because computing the optimal ordering between subtasks is more important in these kinds of subtask graphs. Since only \NSGS{} can take into account the cost of each subtask from the observation, it can find a better sequence of subtasks more often.

In `Base+Distractor', `Base+NOT', and `Base+NegDistractor' cases, it is more important for the agent to carefully find and execute subtasks that have a positive effect in the long run while avoiding distractors that are not helpful for executing future subtasks. In these tasks, the greedy baseline tends to execute distractors very often because it cannot consider the long-term effect of each subtask in principle. On the other hand, our \RProp{} can naturally screen out distractors by getting zero or negative gradient during reward back-propagation. Similarly, \RProp{} performs well on `Base+Delayed' set because it gets non-zero gradients for all subtasks that are connected to the final rewarding subtask. Since our \NSGS{} was distilled from \RProp{}, it can handle delayed reward or distractors as well as (or better than) \RProp{}. 

\begin{table*}[!hbt] 
\centering
\begin{tabular}{r|c|l}
\hline
      & $N_T$       & \{4,3,2,1\}\\ 
      & $N_D$       & \{0,0,0,0\}\\
      & $N_A$       & \{3,3,2\}-\{4,3,3\}\\
\tb{Base}& $N_{ac}^{+}$& \{1,1,2\}-\{3,2,2\}\\ 
      & $N_{ac}^{-}$& \{0,0,0\}-\{0,0,0\}\\
      & $N_{dp}$      &\{0,0,0,0\}-\{0,0,0,0\}\\
      & $N_{oc}$      &\{1,1,1\}-\{2,2,2\}\\
      & $N_{step}$  &40-60\\
  \hline
\tb{-OR}& $N_{oc}$      &\{1,1,1\}-\{1,1,1\}\\
  \hline
\tb{+Distractor}& $N_D$       & \{2,1,0,0\}\\
  \hline
\tb{+NOT}& $N_{ac}^{+}$& \{0,0,0\}-\{3,2,2\}\\
\hline
\tb{+NegDistractor}& $N_D$       & \{2,1,0,0\}\\
& $N_{dp}$      &\{0,0,0,0\}-\{3,3,0,0\}\\
\hline
\tb{+Delayed}& $r$&\{0,0,0,1.6\}-\{0,0,0,1.8\}\\
\hline
\end{tabular}
\vspace{-2pt}
\caption{Subtask graph parameters for analysis of subtask graph components.}
\label{tb:graph_param2}
\end{table*}

\end{document}